\documentclass[11pt]{article}
\usepackage{amsmath,amsfonts,amsmath,amssymb,amsthm}
\usepackage{comment}
\usepackage{bm}
\usepackage{graphicx,psfrag,epsf}
\usepackage{enumerate}
\usepackage{natbib}
\usepackage{algorithm}
\usepackage[noend]{algpseudocode}
\RequirePackage[colorlinks,citecolor=blue,linkcolor=blue,urlcolor=blue]{hyperref}

\usepackage{multirow}
\usepackage{caption}
\usepackage{subcaption}

\newcommand{\blind}{0}

\usepackage{url} 

\usepackage{tikz}
\usepackage[capitalise]{cleveref}

\usepackage[shortlabels]{enumitem}

\newtheorem{theorem}{Theorem}

\newtheorem{corollary}{Corollary}

\newtheorem{proposition}{Proposition}
\newtheorem{definition}{Definition}
\newtheorem{assumption}{Assumption}

\DeclareMathOperator{\Var}{Var}
\DeclareMathOperator{\argmin}{argmin}

\DeclareMathOperator*{\rsa}{\rightsquigarrow}

\newcommand{\Itau}{[\tau_L, \tau_U], \ 0<\tau_L\leq \tau_U < 1}
\newcommand{\n}[2][n]{\binom{#1}{#2}}
\newcommand{\xstar}{\bm{x}^*}

\newcommand{\bmo}{\bm{\omega}}

\newcommand{\SNR}{\text{SNR}}

\newcommand{\Pint}{\mathfrak{P}_{\text{internal}}}
\newcommand{\Ptml}{\mathfrak{P}_{\text{terminal}}}
\newcommand{\NA}{\text{N-A}}

\date{}
\begin{document}

\def\spacingset#1{\renewcommand{\baselinestretch}%
{#1}\small\normalsize} \spacingset{1}

\if0\blind
{
  \title{\bf Global Censored Quantile Random Forest \\}
  \author{\Large{Siyu Zhou and Limin Peng} \\ ~ \\ Department of Biostatistics and Bioinformatics \\ Emory University}

  \maketitle
} \fi

\if1\blind
{
  \bigskip
  \bigskip
  \bigskip
  \begin{center}
    {\LARGE\bf Global Censored Quantile Random Forest}
\end{center}
  \medskip
} \fi

\bigskip
\begin{abstract}
\noindent In recent years, censored quantile regression has enjoyed an increasing popularity for survival analysis while many existing works rely on linearity assumptions.
	In this work, we propose a Global Censored Quantile Random Forest (GCQRF) for predicting a conditional quantile process on data subject to right censoring, a forest-based flexible, competitive method able to capture complex nonlinear relationships. 
	Taking into account the randomness in trees and connecting the proposed method to a randomized incomplete infinite degree U-process (IDUP),
	 we quantify the prediction process' variation without assuming an infinite forest and establish its weak convergence. Moreover, feature importance ranking measures based on out-of-sample predictive accuracy are proposed. We demonstrate the superior predictive accuracy of the proposed method over a number of existing alternatives and illustrate the use of the proposed importance ranking measures on both simulated and real data.
\end{abstract}


\spacingset{1.25}

\section{Introduction}

Survival analysis offers widely-applied, instrumental tools for studying time to events such as death of patients from certain diseases, occurrence of bankruptcy or default, failure of machine components and so on. Such data often exhibits severe skewness and can be incomplete due to censoring or truncation, making the mean time-to-event less insightful or even unidentifiable. In addition, many traditional methods, Cox regression models for example, rely heavily on the proportional hazard assumption that, regardless of the features values, the relative hazard of two subjects remains constant over time, which is hard to justify for real complex applications.

Quantile regression provides a robust, promising alternative to overcome these difficulties and grants the flexibility of focusing on a specific part of the conditional distribution or the entire one. More importantly, it can unravel the heterogeneous effects of features of interest across the entire distribution, offering more valuable and thorough understanding of the problem of interest. Many methods have been proposed to adjust for censoring, including \citep[][among others]{Peng2008, Wang2009}, but often impose a linearity assumption, either globally over a range of probability levels or locally at some probability levels, and utilizes explicit penalization to accommodate high dimensional problems \citep{Zheng2018}. A less restrictive nonparametric approach is the single index model \citep{Christou2019}, which instead assumes the conditional quantile is connected to the linear model through a smooth link function to be estimated by, for example, kernel methods.

Recent advances of machine learning have expanded the toolkit with more flexible and competitive alternatives. Random forest \citep{Breiman2001}, an ensemble of \textit{randomized} trees, is one such model with a well-established record of success across various scientific domains \citep{Fernandez2014}. 
By aggregating randomized trees grown on bootstrap samples each of which partitions the feature space recursively among a set of directions selected at random and fits a simple model in each resulting region,
random forests are flexible enough to model nonlinear complex relationships and adopt naturally to not only high dimensional feature space but also high dimensional responses.

Although random forest is originally proposed for regression and classification problems, such flexibility and exceptional predictive power
inevitably leads to extensions to other areas including survival analysis and quantile regression. In particular, \cite{Li2020} proposed Censored Quantile Regression Forest (CQRF) by designing a weighted estimating equation to adjust for censoring and estimate conditional quantiles where the weights are generated  adaptively from a random forest. However, the random forest utilized in CQRF treats all observed time-to-event as the true values and does not take censoring into consideration. \cite{Zhu2022} proposed Hybrid Censored Quantile Regression Forest which assumes the conditional quantile is a linear model on a set of low dimensional features but the coefficients are modified by another set of, possibly high dimensional, features and estimated based on adaptive weights from a random forest. However, the growing of trees relies on a working linear model, violation of which can result in suboptimal splits. What's more, both forest-based approaches targeted at estimating quantiles locally at a prespecified quantile level.

In this paper, without making any parametric or linear assumptions, we follow closely to Breiman's random forest and propose Global Censored Quantile Random Forest (GCQRF) for predicting the conditional quantile process \emph{globally} over a range of quantile levels on right censored data. As considered in recent theoretical works on random forests \citep[][among others]{Scornet2015, Mentch2016, Wager2018, Peng2022}, subsamples rather than bootstrap samples are utilized for growing trees to avoid duplicated observations. More details of the method will be discussed in section \ref{sec:Model}.

Enjoying advantages of both random forests and quantile regression, our proposed procedure adjusts for censoring while growing trees and is promising with high predictive power, flexible to model complex, nonlinear relationship between response and features of interest, rich of implicit regularization to deal with both high-quality and noisy data, and ready to use for high dimensional settings. 
In addition, our procedure targets on the conditional quantiles, rather than conditional mean, of time-to-event and thus is more robust to skewed data and outliers.
Moreover, predictions and interpretations at two different but close probability levels can exhibit considerable variability due to the limited sample size of real data, despite of lack of scientific evidence of favoring one over the other. Considering this, our objective is to estimate conditional quantiles globally over an interval of quantile levels instead of at a specific value so as to understand the entire conditional quantile process over an interval rather than a particular conditional quantile and to provide a more stable, reliable and complete analysis.

Theoretically, we formulate random forests with process-type predictions as randomized incomplete infinite-degree U-processes (IDUPs), quantify the variance of randomized incomplete IDUPs and establish the corresponding functional central limit theorem (FCLT). To the best of our knowledge, our result is the first one to quantify the variance of random forests with process-type predictions  while taking the random-feature-subsetting into consideration explicitly without assuming the forest consists of trees growing on all possible subsamples.
Similar results on the original random forests date back to \cite{Mentch2016} which first made the connection between random forests and U-statistics and established the asymptotic normality of random forests for real-valued response. \cite{Peng2022} further relaxed conditions for asymptotic normality and provided Berry-Essen bounds to characterize the rate of convergence. Results in \cite{Mentch2016} and \cite{Peng2022} focused on the variation of random forest predictions around the expected tree predictions without quantifying the bias or showing consistency. Whether such unbiasedness is beneficial can be application-dependent, especially when the accuracy is the primary target.

Additional assumptions or specifications are generally needed to establish the consistency of random forests. \cite{Scornet2015} derived the first consistency result on Breiman's original random forest with the assumption that data is generated from an additive model. \cite{Wager2018} required specific settings of tree structure, including honesty, $\alpha$-regularity and random-splitting, to guarantee consistency and dominance of the variance of random forests over squared bias and established asymptotic unbiasedness and normality. Interested readers are referred to \cite{Wager2018} for details of these specifications. Following the perspective of viewing random forests as adaptive nearest neighbor estimators \citep{Hothorn2004, Meinshausen2006}, \cite{Athey2019} generalized the results in \cite{Wager2018} and proposed the generalized random forest framework to estimate parameters in the Euclidean space that can be identified by estimating equations, including but not limited to the conditional mean and the conditional quantiles. Thanks to \cite{Wager2018} and \cite{Athey2019}, there have been a growing literature to develop tree-based methods for various problem settings \citep[][among others]{Li2020, Cho2023, Cho2022}.

However, there are two limitations in the asymptotic results in \cite{Wager2018} and those with similar settings. First, the tree under consideration is one with the randomness from feature subsetting marginalized out, and thus can be considered as a \emph{fixed} kernel. However, recent work \citep{Mentch2020} suggests the random-feature-subsetting play a key role in the success of random forests by providing an implicit regularization effect so that random forests are adaptive to data of various quality. Moreover, this implicit regularization effect is not limited to trees but also applicable to linear models, ensembles of which can, surprisingly, outperform explicitly regularized procedures such as lasso even when the data is generated from a sparse linear model. Thus, it is beneficial to take the randomness into account for both theoretical and practical purposes. 
Moreover, the random forest under consideration is a nearly infinite one in the sense that it consists of all trees grown on all possible subsamples. However, with subsample size depending on the training size $n$, the number of all possible subsamples grows dramatically and it quickly becomes computationally infeasible to grow such infinite forests. Practically, only a randomly selected set of subsamples are utilized for growing the forest, leading to the limiting variance larger than, possibly dominating, the variance in the infinite case. Thus, it is crucial to take this finiteness into account.

Our work seeks to fill in the gap that existing theoretical studies for asymptotic Gaussianty of random forests are not applicable to random forests with infinite-dimensional process-type predictions. A related work towards this direction is \cite{Heilig2001} which studied the uniform weak law of large number (UWLLN) and functional central limit theorem (FCLT) of U-processes of fixed kernels with degrees growing with $n$, but results in \cite{Heilig2001} are not directly applicable for randomized trees or when not all subsamples are used for growing the forest. Our results follow and extend those in \cite{Peng2022} and establish the weak convergence of a randomized incomplete IDUP under regularity conditions. As in \cite{Mentch2016} and \cite{Peng2008}, we focus on the convergence around the mean tree prediction and do not impose additional requirements to guarantee consistency. Research along this direction is left as future work.

The rest of the paper is organized as follows. In section \ref{sec:Model}, we present the details of the proposed GCQRF. With preliminaries on the connections of GCQRFs and randomized incomplete IDUPs as well as extended H-decomposition of U-processes of randomized kernels discussed in section \ref{sec:Prelim}, weak convergence of randomized IDUPs are established in Sec \ref{sec:Theory}. We propose feature importance ranking measures with respect to conditional quantiles of pre-specified quantile levels in section \ref{sec:vimp}. Section \ref{sec:num} is devoted to numerical analysis on predictive accuracy and feature importance ranking measures of GCQRFs with both simulated and real data. We conclude with additional discussions in section \ref{sec:discussions}.

\section{Global Censored Quantile Random Forest}\label{sec:Model}

Suppose we have a random sample $\mathbb{D}_n =\{ \bm{Z}_i = (\bm{X}_i, O_i): \ i = 1, \dots, n\}$ i.i.d. following $F_{\bm{X}, O}$ with features $\bm{X} = (X_1, \dots, X_p) \in \mathcal{X}$ and the response of interest $O \in\mathcal{O}$. For right censored data, $O = (Y, \delta)$ where $Y = \min(T, C)$ is the observed response and $\delta = I(T \leq C)$ is the censoring indicator with $T$ the true underlying response and $C$ the censoring. Without loss of generality, we assume $Y$, $C$, and $T$ are all real-valued. 

Given a new observation at $\bm{X} = \bm{x}$, our GCQRF aims to predict the associated conditional quantile process
\begin{align*}
	Q_{T}(\tau |\bm{X} = \bm{x}) = \inf \{t: \Pr(T \leq t | \bm{X} = \bm{x} \}, \quad, \tau \in I_{\tau} = \Itau,
\end{align*}
and follows closely to the original random forest for classification and regression. More specifically, draw $B$ subsamples, each of size $s_n$ randomly selected without replacement, and grow a tree on each by recursively partitioning the feature space with each split perpendicular to some axis (feature) $X_j$ among \texttt{mtry} features selected uniformly at random at some cutting position so as to optimize some criterion. In each region, also called node, $\mathcal{N}$, estimate the conditional quantile process based on observations in the node. Repeat this until the number of nodes reaches a prespecified value \texttt{maxnodes} , or the number of uncensored observations in each terminal node, called leaf, is less than a prespecified value \texttt{nodesize}. In addition, a split will not be considered if the number of unique uncensored observations in either child node is less than a prespecified value \texttt{nodesizeMin}. 

It remains to specify the splitting criterion and estimations in each node. We propose a splitting rule based on a modified version of the loss function for quantile regression for censored data proposed in \cite{Li2017}. More specifically, we consider the integrated quantile loss (I-QLoss) defined by
\begin{align*}
	L_{\rho}\left (\widehat{Q}_{T}(\tau |\bm{X})\right ) = \int_{\tau_L}^{\tau_U} L_{\rho}\left (\tau, \widehat{Q}_{T}(\tau |\bm{X})\right ) d\tau
\end{align*}
where $\widehat{Q}_{T}(\tau|\bm{X})$, $\tau \in [\tau_L, \tau_U]$, is the estimated conditional quantile process and
\begin{align*}
	L_{\rho}\left (\tau, \widehat{Q}_{T}(\tau |\bm{X})\right ) = \mathbb{P}\left [\tau(1-\tau) \rho_{\tau}\left (T^{u} - \widehat{Q}_{T}(\tau|\bm{X})\right )\right ],
\end{align*}
is the quantile loss at $\tau$ with $\rho_{\tau}(v) = v(\tau - I(v < 0))$, $T^{u} = \min(T, u)$ and $u$ a deterministic prespecified constant less than the upper bound of $C$'s support. For brevity, we simplify $Q_{T}(\tau |\bm{X})$ and $\widehat{Q}_{T}(\tau |\bm{X})$ by $Q(\tau |\bm{X})$ and $\widehat{Q}(\tau |\bm{X})$ in the rest of this article when there is no ambiguity.  In $L_{\rho}\left (\widehat{Q}_{T}(\tau |\bm{X})\right )$ as well as notations below depending on the probability interval $[\tau_L, \tau_U]$ of interest, we omit this the dependence for readability.

Given a node $\mathcal{N}$ containing censored observations, QLoss at $\tau$ is estimated by 
\begin{align*}
	\widehat{L}_\rho\left (\tau, \widehat{Q}(\tau |\bm{X} \in \mathcal{N})\right ) =   \frac{1}{n_{\mathcal{N}}} \sum_{i \in \mathcal{N}}  \frac{\Delta_i}{\widehat{G}(Y^u_i  |\bm{X} \in \mathcal{N})} \tau(1-\tau) \rho_{\tau}\left (Y_i^u - \widehat{Q}(\tau |\bm{X} \in \mathcal{N})\right )
\end{align*}
where $\{i \in \mathcal{N}\}$ is short for $ \{i: \bm{X}_i \in \mathcal{N} \}$, $\widehat{Q}(\tau |\bm{X} \in \mathcal{N})$ is the predicted $\tau$-th quantile in node $\mathcal{N}$, $n_{\mathcal{N}}$ is the number of observations in $\mathcal{N}$, $Y^u_i = T^u_i \wedge C_i = Y_i \wedge u$, $\Delta_i = I(T^u_i \leq C_i) = I(Y_i >u) + I(Y_i \leq u, \delta_i = 1)$ and $\widehat{G}(t|\bm{X} \in \mathcal{N})$ is the Nelson-Aalen estimator using observations in node $\mathcal{N}$ for $Pr(C>t|\bm{X} \in \mathcal{N})$. On the other hand, in case of a node without any censored observations, estimate QLoss by
\begin{align*}
	\widehat{L}_\rho\left (\tau, \widehat{Q}(\tau |\bm{X} \in \mathcal{N})\right ) =   \frac{1}{n_{\mathcal{N}}} \sum_{i \in \mathcal{N}}   \tau(1-\tau) \rho_{\tau}\left (Y_i - \widehat{Q}(\tau |\bm{X} \in \mathcal{N})\right ).
\end{align*}
The I-QLoss over $[\tau_L, \tau_U]$ is then estimated by
\begin{align*}
	\widehat{L}_{\rho}(\mathcal{N}):= \int_{\tau_L}^{\tau_U}\widehat{L}_\rho\left (\tau, \widehat{Q}(\tau |\bm{X} \in \mathcal{N})\right ) d\tau.
\end{align*}

Given a parent node $\mathcal{P}$ to be split, let $C_1$ and $C_2$ denote the left and right child node respectively, and we propose to find the optimal split among randomly selected candidate features that maximizes
\begin{align*}
	n_{\mathcal{P}}\widehat{L}_{\rho}(\mathcal{P}) - n_{C_1}\widehat{L}_{\rho}(C_1) - n_{C_2}\widehat{L}_{\rho}(C_2).
\end{align*}

What is left is how to estimate $\widehat{Q}(\tau |\bm{X} \in \mathcal{N})$ for any node $\mathcal{N}$. To achieve this, we first construct a Nelson-Aalen estimator $\widehat{H}(t|\bm{X} \in \mathcal{N})$ for the survival distribution of $T$ based on observations in $\mathcal{N}$. Let $\tau_c = \sup\{t: \widehat{H}(t|\bm{X}\in \mathcal{N}) > 0 \}$. For $\tau \leq \tau_c$, take $\widehat{Q}(\tau |\bm{X} \in N) = \inf\{t: \widehat{H}(t|\bm{X}\in \mathcal{N})\leq 1-\tau \}$. For $\tau > \tau_c$, take $\widehat{Q}(\tau |\bm{X} \in N)$ as the largest observation in $\mathcal{N}$ or the largest uncensored observation in $\mathcal{N}$ or extrapolating the estimated Nelson-Aalen estimator with exponential distribution and inverting it to get the estimated quantiles.

As in the original regression tree, the $b$-th tree prediction $\widehat{Q}^{b}(\tau|\bm{X} = \bm{x})$, $b = 1, \dots, B$, $\tau \in [\tau_{L}, \tau_{U}]$, at any point $\bm{x}$ is the prediction $\widehat{Q}(\tau |\bm{X} \in \mathcal{N})$ where $\mathcal{N}$ is the node that $\bm{x}$ falls in, and the forest prediction is the average of $B$ trees' predictions
\begin{align*}
	\widehat{Q}(\tau|\bm{X} = \bm{x}) = \frac{1}{B} \sum_{b=1}^{B} \widehat{Q}^{b}(\tau|\bm{X} = \bm{x}).
\end{align*}
In practice, we find the predicted conditional quantiles on a fine grid $\mathcal{S}_{\tau} = \{0 < \tau_L = \tau_1 < \dots < \tau_{n_\tau} = \tau_{U} < 1 \}$ and the predicted quantile process is a step function with jumps at $\tau \in \mathcal{S}_{\tau}$.

Following closely to the Algorithm 1 in \cite{Biau2016}, Algorithm \ref{alg:GCQRF} provides a summary for GCQRFs with stopping rule based on \texttt{nodesize}. To control the tree size by \texttt{maxnodes} on top of \texttt{nodesize}, simply replace the condition in the while loop by comparing the size of $\Ptml$ to \texttt{maxnodes}.

It is noticeable that the proposed GCQRF can also work on complete data without any censoring. In contrast to existing forests for quantile regression including \cite{Meinshausen2006} using the CART criterion and \cite{Athey2019} utilizing multiclass classification rule on relabeled responses, the proposed splitting rule minimizes the sum of I-QLoss in the two children nodes with respect to the standard quantile loss $\rho_{\tau}(v)$.

\begin{algorithm}[!ht]
	\caption{Global Cencored Quantile Random Forests (GCQRFs)}
	\label{alg:GCQRF}
	\begin{algorithmic}[1]
	\Procedure{GCQRF}{Targeted location $\bm{x}$, a $\tau$-grid $\mathcal{S}_{\tau} = \{0  < \tau_1 = \tau_L < \dots < \tau_{n_\tau} = \tau_{U} < 1 \}$, training sample $\mathbb{D}_n = \{(\bm{X}_i, Y_i, \delta_i): i = 1, \dots, n\}$, number of trees \texttt{ntree}, subsampling size $s_n$, number of candidate features \texttt{mtry}, respective upper and lower bound for the number of uncensored observations in leaves \texttt{nodesize} and \texttt{nodesizeMin}}
		\For{$b = 1,\dots,B $}  
			\State Draw a subsample $\mathcal{D}_{b}$ with subsampling size $s_n$ that contains at least one uncensored observation.
			\State Let $\Pint = \{\mathcal{X}\}$ be the set containing nodes to be split.
			\State Let $\Ptml = \emptyset$ be the set containing leaves.
			\While{$\Pint \neq \emptyset$}
				\State Let $\mathcal{P}$ denote the first element in $\Pint$.
				\If{$\mathcal{P}$ contains less than \texttt{nodesize} uncensored observations or if all $\bm{X}_i \in \mathcal{P}$ are equal }	
					\State Move $\mathcal{P}$ from $\Pint$ to $\Ptml$.
				\Else
					\State Select \texttt{mtry} features as candidates uniformly at random without replacement.
					\State Among all possible splits in the candidates such that there are at least \texttt{nodesizeMin} unique
uncensored observations in daughter nodes $\mathcal{L}$ and $\mathcal{R}$, find the optimal one maximizing
					\begin{align*}
						n_{\mathcal{P}}\widehat{L}_{\rho}(\mathcal{P}) - n_{C_1}\widehat{L}_{\rho}(\mathcal{L}) - n_{C_2}\widehat{L}_{\rho}(\mathcal{R}).
					\end{align*}
					\State Partition $\mathcal{P}$ based on the optimal split. 
					\State Obtain Nelson-Aalen estimators in $\mathcal{L}$ and $\mathcal{R}$ respectively.
					\State Add $\mathcal{L}$ and $\mathcal{R}$ to $\Pint$.
				\EndIf
			\EndWhile
			\State Find the node $\mathcal{N}$ in $\Ptml$ that $\bm{x}$ falls.
			\State Compute the predicted quantiles of the tree $\widehat{Q}^{b}(\tau_{k}|\bm{X} = \bm{x}), k = 1, \dots, n_{\tau}$, based on the Nelson-Aalen estimator associated with $\mathcal{N}$.
		\EndFor 
		\State Compute the predicted quantiles of the forests by averaging trees' predictions
		\begin{align*}
			\widehat{Q}(\tau_{k}|\bm{X} = \bm{x}) = \frac{1}{B} \sum_{b=1}^{B} \widehat{Q}^{b}(\tau_{k}|\bm{X} = \bm{x}), \quad k = 1, \dots, n_{\tau}.
		\end{align*}
		
	\EndProcedure
	\end{algorithmic}
\end{algorithm}

\section{Preliminaries}\label{sec:Prelim}

\cite{Mentch2016} connected random forests to an incomplete infinity degree U-statistic (IDUS) of randomized kernels, and established the asymptotic normality for random forest predictions. \cite{Peng2022} further extended U statistic and the Hoeffding-decomposition (H-decomposition) to a more generalized notion which put complete or incomplete U statistics of fixed or randomized kernels under the same umbrella and established more relaxed conditions for the asymptotic normality of random forests and a Berry-Essen bound to provide a more refined characterization of the rate of convergence. 

In this section, we will first demonstrate that GCQRF can be similarly viewed as a variant of infinity degree U-process (IDUP) which will be shown to converge weakly to a Gaussian process in the next section, and then extend the generalized H-decomposition of U-statistic to the case of U-process. 

\subsection{Global Censored Quantile Random Forests as Randomized Incomplete Infinite Degree U-Processes}\label{sec:Prelim_UProcess}

Let $\mathcal{H}_{s} = \{h_{s}(z_1,\dots, z_{s};\tau): \tau \in I_\tau \}$ be a class of permutation-symmetric kernels of degree $s$ indexed by $\tau$. Let $\theta_{s}(\tau)$ be the mean of the kernel $\mathbb{P} h_{s}(\bm{Z}_1, \dots,  \bm{Z}_{s}; \tau)$.
Denote $(n,s) = \{\beta = (\beta_1, \dots, \beta_s):1 \leq \beta_1 < \dots < \beta_s \leq n\}$ the set of all ordered $s$-tuples from $\{1, \dots, n\}$ and $(n)_{s} = \{\beta = (\beta_1, \dots, \beta_s):1 \leq \beta_1, \dots, \beta_s \leq n\}$ the set of unordered ones. For each $\beta \in (n,s)$, denote $\mathcal{D}_{s, \beta} = \{\bm{Z}_{\beta_1}, \dots, \bm{Z}_{\beta_s}\} $ a subsample of $\mathbb{D}_n$ of size $s$.

Define 
\begin{align*}
	U_{n, s}h_{s}(\tau) 
	= \frac{1}{\n{s}} \sum_{\beta \in (n, s)} h_{s}(\mathcal{D}_{s,\beta}; \tau)
	= \frac{1}{\n{s}} \sum_{\beta \in (n, s)} h_{s}(\bm{Z}_{\beta_1}, \dots,  \bm{Z}_{\beta_{s}}; \tau),
\end{align*}
which, for each $\tau$, is a U-statistic of degree $s$ with kernel $h_{s}(z_1,\dots, z_{s}; \tau)$,
and $\{U_{n, s}h_{s}(\tau): h_{s}(z_1, \dots, z_{s};\tau) \in \mathcal{H}_{s} \}$ be the corresponding U process. For simplicity, we drop $s$ in $\mathcal{D}_{s, \beta}$ since this is reflected in the degree of the kernel $h_s$.


First let us consider a \textit{fixed} tree grown on a subsample $\mathcal{D}_{\beta}$ without any additional randomness whose prediction of $\tau$-th quantile at $\bm{x}^*$ is denoted by $h_s(\mathcal{D}_{\beta};\tau, \xstar)$ with $\tau \in I_\tau = \Itau$. This kind of tree can be achieved by considering all available features at each split. Another example is an \textit{pseudo average} tree defined as the expectation of a randomized tree over the randomness \citep{Wager2018}.

Grow an ensemble of $\n{s}$ trees on all the possible $\n{s}$ subsamples of size $s$ and take the average of their predictions at $\xstar$, leading to the subbagged predictions at $\xstar$
\begin{align}
	U_{n, s}h_{s}(\tau;\bm{x}^{*}) =  \frac{1}{\n{s}} \sum_{\beta \in (n,s)} h_{s}(\mathcal{D}_{\beta};\tau, \bm{x}^{*}) \label{eqn:bag}
\end{align}
Since tree predictions are invariant to the order of observations in a subsample, for each $\tau$, $U_{n, s}h_{s}(\tau;\bm{x}^{*})$ is in the form of a U-statistic with the tree being a kernel of degree $s$. Then $\{U_{n, s}h_{s}(\tau;\bm{x}^{*}): \tau \in \Itau \}$ is a U-process of degree $s$ indexed by $\tau$.

Notice that the subsample size $s$ in (\ref{eqn:bag}) is a fixed value, which limits how deep a tree can be grown. To make the model more flexible for potentially improved performance, let $s$ grow with $n$, giving rise to 
\begin{align}
	U_{n, s_n}h_{s_n}(\tau;\bm{x}^{*}) = \frac{1}{\n{s_n}} \sum_{\beta \in (n,s_n)} h_{s_n}(\mathcal{D}_{\beta};\tau, \bm{x}^{*}). \label{eqn:bag_kn}
\end{align}
The resulting process $\{U_{n, s_n}h_{s_n}(\tau;\bm{x}^{*}): \tau \in I_{\tau} = \Itau\}$ is an IDUP \citep{Heilig2001}. 

Since $\n{s_n}$ grows fast with $n$, it quickly becomes computational infeasible if all $\n{s_n}$ subsamples are considered. One way of overcoming this is to consider a bootstrap of $m_n$ subsamples randomly selected with replacement from all $\n{s_n}$ possible ones\citep{Mentch2016, Chen2019}, which is also the bootstrapping scheme considered in our proposed GCQRF in Algorithm \ref{alg:GCQRF}. With a little abuse of notation, let $\bm{M}_{n, s_n} = \left(M_1, \dots, M_{\n{s_n}}\right ) \sim Multinomial\left (m_n; \frac{1}{\n{s_n}}, \dots, \frac{1}{\n{s_n}}\right )$ where, for $\beta = 1, \dots, \n{s_n}$, $M_{\beta}$ denotes the number of times $\mathcal{D}_{\beta}$, $\beta \in (n, s_n)$, is selected. This leads to the \textit{incomplete} IDUP 
\begin{align}
	U_{n, s_n, m_n}h_{s_n}(\tau;\bm{x}^{*}) = \frac{1}{m_n} \sum_{\beta \in (n,s_n)} M_{\beta} h_{s_n}(\mathcal{D}_{\beta};\tau, \bm{x}^{*}), \ \tau \in [\tau_L, \tau_U] \label{eqn:bag_kn_mn}.
\end{align}

An alternative way of reducing computational cost is to select the subsamples based on $\bm{\rho}_{n, s_n} = \left(\rho_1, \dots, \rho_{\n{s_n}}\right)$, a vector of i.i.d. Bernoulli random variables with $\Pr(\rho_\beta=1) := p_n = N/\n{s_n} $ the probability that $\mathcal{D}_{\beta}$ is selected \citep{Chen2019, Peng2022}. The total number of subsample selected is equal to $\hat{N} = \sum_{\beta \in (n,s_n)}\rho_{\beta}$ where $\mathbb{P}(\hat{N}) = N$. When $p_n < 1$, the incomplete IDUP bootstrapped in this way can be written as
\begin{align}
	U_{n, s_n, N}h_{s_n}(\tau;\bm{x}^{*}) = \frac{1}{\hat{N}} \sum_{\beta \in (n,s_n)} \rho_{\beta} h_{s_n}(\mathcal{D}_{\beta};\tau, \bm{x}^{*}), \ \tau \in [\tau_L, \tau_U] \label{eqn:bag_kn_N}.
\end{align}
When $p_n = 1$, i.e., $N = \n{s_n}$, it reduces to the complete IDUP.

Recall that trees in the proposed GCQRF are randomized such that only a subset of features are randomly selected at each split as candidates and such selection is independent of the samples used to grow a tree and is independent across trees. Denoting $\bmo = \left(\omega_1, \dots, \omega_{\n{s_n}}\right)$ and $h_{s_n}(\mathcal{D}_{|beta}, \omega_{\beta};\tau, \bm{x}^{*})$ the $\beta$-th tree, our proposed GCQRF as in Algorithm \ref{alg:GCQRF} is a randomized incomplete IDUP grown on $m_n$ subsamples randomly selected with replacement from all $\n{s_n}$ possible subsamples and its $\tau$-th quantile prediction at $\xstar$ is defined by
\begin{align}
	U_{ n, s_n, m_n}h_{s_n}(\omega, \tau;\bm{x}^{*}) = \frac{1}{m_n} \sum_{\beta \in (n, s_n)} M_{\beta} h_{s_n} (\mathcal{D}_{\beta}, \omega_{\beta};\tau, \bm{x}^{*}). \label{eqn:RandomizedIncompleteIDUP_M}
\end{align}
Though not implemented in Algorithm \ref{alg:GCQRF}, GCQRF can be alternatively grown based on $\bm{\rho}_{n,s_n}$ and the prediction is then 
\begin{align}
	U_{n, s_n, N}h_{s_n}(\omega, \tau;\bm{x}^{*}) = \frac{1}{\hat{N}} \sum_{\beta \in (n, s_n)} \rho_{\beta} h_{s_n} (\mathcal{D}_{\beta}, \omega_{\beta};\tau, \bm{x}^{*}). \label{eqn:RandomizedIncompleteIDUP_rho}
\end{align}
When each unique subsamples of size $s_n$ is selected once, then it becomes a randomized complete IDUP
\begin{align}
	U_{n, s_n}h_{s_n}(\omega, \tau;\bm{x}^{*}) = \frac{1}{\n{s_n}} \sum_{\beta \in (n, s_n)} h_{s_n} (\mathcal{D}_{\beta}, \omega_{\beta};\tau, \bm{x}^{*}). \label{eqn:RandomizedCompleteIDUP}
\end{align}
In the following, we omit the dependence on $\xstar$ when there is no ambiguity.

\subsection{Extended Hoeffiding Decomposition of Randomized U-Processes}

A useful and widely applied technique in the U-statistic or U-process literature is Hoeffiding decomposition (H-decomposition) where the components are themselves U-statistics or U-processes and, more importantly, uncorrelated with variance of decreasing order \citep{uStat, Heilig2001}. Recall that $s$ is the degree of the kernel and, in the rest of this subsection, can be either fixed or grow with $n$.


\cite{Peng2022} extended the H-decomposition of U-statistic of fixed kernels to the case of randomized kernels, which can naturally be extended to U-processes. For $j=1, \dots, s-1$ and for each $\tau$, define the conditional kernel by
\begin{align*}
	h_{s|j}(\bm{z}_1, \dots, \bm{z}_j;\tau) = \mathbb{P}\left [ h_{s}(\bm{Z}_1, \dots, \bm{Z}_{s}, \omega;\tau) | \bm{Z}_1 = \bm{z}_1, \dots, \bm{Z}_j = \bm{z}_j \right ]
\end{align*}
where the expectation is with respect to $\bm{Z}_{j+1}, \dots, \bm{Z}_s$ and $\omega$, and define the projected kernel by
\begin{align*}
	h_{s}^{(0)}(\tau) = \theta_{s}(\tau),	\\
	h_{s}^{(j)}(\bm{z}_1, \dots, \bm{z}_j;\tau) &= h_{s|j}(\bm{z}_1, \dots, \bm{z}_j;\tau) - \sum_{k=0}^{j-1}\sum_{(j, k)} h_{s}^{(k)}(\bm{z}_{i_1}, \dots, \bm{z}_{i_{k}};\tau),	\\
	h_{s}^{(s)}(\bm{z}_1, \dots, \bm{z}_s, \omega; \tau) &= h_{s}(\bm{z}_1, \dots, \bm{z}_s, \omega;\tau) - \sum_{k=0}^{j-1}\sum_{(j, k)} h_{s}^{(k)}(\bm{z}_{i_1}, \dots, \bm{z}_{i_{k}};\tau).
\end{align*}

For $j = 1, \dots, s-1$ and $\tau, \tau' \in [\tau_L, \tau_U]$, define
\begin{align}
	\zeta_{j,s}(\tau, \tau') &=  \mathbb{P} \left[h_{s|j}(\bm{Z}_1, \dots, \bm{Z}_j;\tau) - \theta_{s}(\tau))\right] \left[h_{s|j}(\bm{Z}_1, \dots, \bm{Z}_j;\tau') - \theta_{s}(\tau'))\right],	\label{eqn:Var_ConditionalKernel}	\\
	\zeta_{s,s}(\tau, \tau') &=  \mathbb{P} \left[h_{s}(\bm{Z}_1, \dots, \bm{Z}_s, \omega;\tau) - \theta_{s}(\tau))\right] \left[h_{s}(\bm{Z}_1, \dots, \bm{Z}_s, \omega;\tau') - \theta_{s}(\tau'))\right]. \label{eqn:Var_TreeKernel}
\end{align}


The extended H-decomposition of a randomized U-process $U_{n, s}h_{s}(\omega, \tau)$ is given by
\begin{align}
	U_{n, s}h_{s}(\omega, \tau) &= \sum_{j=0}^{s-1} \n[s]{j} \n{j}^{-1} \sum_{(n,j)} h_{s}^{(j)}(\bm{Z}_{i_1}, \dots, \bm{Z}_{i_j}; \tau) +  \n{s}^{-1} \sum_{(n,s)} h_{s}^{(s)}(\bm{Z}_{i_1}, \dots, \bm{Z}_{i_s}, \omega; \tau) \nonumber \\
			&= \sum_{j=0}^{s-1} \n[s]{j} U_{n, j}h_{s}^{(j)}(\tau) +   U_{n, s}h_{s}^{(s)}(\omega, \tau).\label{eqn:HDecomp}	
\end{align}
In particular the first order term is given by
\begin{align*}
	sU_{n,1}h_{s}^{(1)}(\tau) = \frac{1}{n}\sum_{i=1}^{n}s\left [ h_{s|1}(\bm{Z}_i;\tau) - \theta_{s}(\tau) \right ],
\end{align*}
a sum of i.i.d processes. Without the additional randomness $\omega$, it reduces to the traditional H-decomposition. Following \cite{Heilig1997}, we call a kernel degenerate if it is of mean-zero.

\section{Theoretical Studies}\label{sec:Theory}

\subsection{Asymptotics of Randomized Complete IDUPs}

In this section, we extend results in \cite{Heilig2001} by accommodating randomized kernels and incompleteness and allowing that the convergence rate depends on the kernel and is not necessarily $\sqrt{n}$, and study the asymptotic properties of randomized complete or incomplete IDUPs of a generic kernel under regularity conditions, which can be applied more broadly beyond tree-based methods.

Through out this section, we assume the tree kernels $\mathcal{H}_{s_n} = \{h_{s_n}(\omega, \tau)\} \subset l^{\infty}(I_{\tau})$ , the space of uniformly bounded real functions with index $\tau \in I_{\tau} = [\tau_L, \tau_U]$. As in \cite{Heilig2001}, all arguments in this section and related proofs apply directly when the supremum is taken over a countable class of events and can be generalized to uncountable classes following discussions on measurability issues in \cite{Pollard1984}.

\begin{assumption}[Randomness] \label{asmp:Randomness}
	For randomized complete IDUP defined in (\ref{eqn:RandomizedCompleteIDUP}) or randomized incomplete IDUPs defined in (\ref{eqn:RandomizedIncompleteIDUP_M}) or (\ref{eqn:RandomizedIncompleteIDUP_rho}), the additional randomness $\omega_{\beta}$, $\beta = 1, \dots, \n{s_n}$, are i.i.d., and independent of the original training sample $\mathbb{D}_{n}$ as well as the selection of subsamples according to
	\[\bm{M}_{n, s_n} = \left (M_1, \dots, M_{\n{s_n}}\right) \quad \text{ or } \quad \bm{\rho}_{n} = \left (\rho_1, \dots, \rho_{\n{s_n}}\right ).\]
\end{assumption}

Weak convergence of a randomized complete IDUP can be established when the first order term $s_nU_{n,1}h_{s_n}^{(1)}(\tau) = \frac{1}{n}\sum_{i=1}^{n}s_n\left [ h_{s_n|1}(\bm{Z}_i;\tau) - \theta_{s_n}(\tau) \right ]$ in the extended H-decomposition converges weakly and dominates the remainders.
With Assumption \ref{asmp:Randomness}, results in \cite{Heilig2001} for controlling the the remainders in the H-decomposition of a complete IDUP of fixed kernel can be generalized to the case of randomized kernels. Details can be found in the supplement material. The order of the remainders depend extensively on the complexity and envelope of the projected kernels, but can be well controlled when the kernel $h_{s_n}$ is uniformly bounded and Euclidean. A wide range of kernels are indeed Euclidean, including trees in the GCQRF and the VC class.

\begin{definition}[Covering Number]
	For a class $\mathcal{G}$ with pseudometric $d$, for each $\epsilon > 0$, the covering number $N(\epsilon, d, \mathcal{G})$ is the smallest integer $N$ for which there exist points $g_1, \dots, g_N$ such that $\min_{i \leq N} d(g, g_i) \leq \epsilon$ for every $g \in \mathcal{G}$.
\end{definition}

\begin{definition}[Euclidean Function Class] 
	Let $G$ be the envelop of the class $\mathcal{G}$. $G$ is Euclidean$(A, V)$ for the envelop $G$ if there exist constants $A$ and $V$ such that for all $\epsilon \in (0, 1]$ and all measures $Q$, the covering number $N(\epsilon, d_{Q, G}, \mathcal{G}) \leq A \epsilon^{-V}$ where $d_{Q, G} = [Q(|g-g'|^2)/Q(G^2)]^{1/2}$.
\end{definition}

\begin{assumption}[Euclidean]\label{asmp:Euclidean}
	The randomized kernel $h_{s_n}(\omega)$ is Euclidean $(A, V)$ with constants $A$ and $V$ for some envelope $H_{s_n}$.
\end{assumption}

Weak convergence of the first order term $s_nU_{n,1}h_{s_n}^{(1)}(\tau) = \frac{1}{n}\sum_{i=1}^{n}s_n\left [ h_{s_n|1}(\bm{Z}_i;\tau) - \theta_{s_n}(\tau) \right ]$ can be established by appealing to empirical process theory on function classes changing with $n$ \citep{Pollard1990}. \cite{Heilig2001} considered the case when
 $s_n^2 \zeta_{1, s_n}(\tau, \tau')$ converges for each $\tau, \tau'$ and thus the converge rate is $\sqrt{n}$. However, this can fail for kernels such as the nonadaptive trees where the tree structure and tree predictions depend on two disjoint training set \citep{Wager2018, Peng2022}. Assumption \ref{asmp:Var_1stOrder_SimilarMagnitude} and \ref{asmp:Cor_1stOrderProjection} below are utilized to accommodate such nonadaptive treees and better characterize the convergence rate. In addition, Assumption \ref{asmp:Envelope_FiniteSecondMoment} guarantees a well-behaved envelope of the first order projected kernel $h_{s_n}^{(1)}(\tau)$.
 
Define
\begin{align}
	r_{1n}^2 &= 	
			\sup_{\tau \in I_{\tau}} \zeta_{1,s_n}(\tau, \tau), 	\quad
	\rho_{1n}^2(\tau, \tau') = 
		\frac{\zeta_{1,s_n}(\tau, \tau) + \zeta_{1,s_n}(\tau', \tau') - 2\zeta_{1,s_n}(\tau, \tau')}{r_{1n}^{2}}.	\label{eqn:ConvRate}
\end{align}

\begin{assumption} \label{asmp:Var_1stOrder_SimilarMagnitude}
	There exists $d_{1}(\tau) \geq C_1 >0 $ for some constant $C_1$ s.t. $\frac{\zeta_{1,s_n}(\tau, \tau)}{r_{1n}^{2}} $ converges uniformly to $ d_1^2(\tau)$ for $\tau \in [\tau_L, \tau_U]$.
\end{assumption}

\begin{assumption} \label{asmp:Cor_1stOrderProjection}
	There exists $H_{1}(\tau, \tau')$ s.t. $\frac{\zeta_{1,s_n}(\tau,\tau')}{\sqrt{\zeta_{1,s_n}(\tau,\tau)\zeta_{1,s_n}(\tau',\tau')}}$ converges uniformly to $H_1(\tau,\tau')$ for $\tau, \tau' \in [\tau_L, \tau_U]$.
\end{assumption}

\begin{assumption} \label{asmp:Envelope_FiniteSecondMoment}	\
	\begin{align*}
		\lim_{n} \frac{\mathbb{E}\left[ \sup_{\tau} (h_{s_n|1}(\bm{Z}_i;\tau) - \theta_{s_n}(\tau))^2 \right]}{r_{1n}^{2}} = 
		\lim_{n} \frac{\mathbb{E}\left[ \sup_{\tau} (h_{s_n|1}(\bm{Z}_i;\tau) - \theta_{s_n}(\tau))^2 \right]}{\sup_{\tau} \mathbb{E}\left[\left(h_{s_n|1}(\bm{Z}_i;\tau) - \theta_{s_n}(\tau)\right)^2 \right]} < \infty
	\end{align*}
\end{assumption}

Lastly,  Assumption \ref{asmp:TreeBehavior_SubRate} on the subsampling rate $s_n$ guarantees that $s_nU_{n,1}h_{s_n}^{(1)}(\tau)$ converges at a rate dominating the remainders. For nonadaptive trees, $s_n = n^{\beta_{1}}$ for $\beta_{1} < 1/3$ would be sufficient.

\begin{assumption}[Tree Behavior and Subsampling Rate] \label{asmp:TreeBehavior_SubRate}
	Assume $s_n= o(\sqrt{nr_{1n}^{2}})$.
\end{assumption}

Define
\begin{align}
	\rho_{1}^2(\tau, \tau') &=
			d_{1}^{2}(\tau) + d_{1}^{2}(\tau') - 2H_1(\tau, \tau') d_{1}(\tau)d_{1}(\tau'), \quad
	\sigma_{1}(\tau, \tau')	=
			H_{1}(\tau, \tau')d_{1}(\tau)d_{1}(\tau').	\label{eqn:Cor_1stOrder}
\end{align}

Putting these together leads to the weak convergence of the randomized complete IDUPs.
\begin{theorem}[Weak Convergence of Randomized Complete IDUPs]\label{thm:WeakCov_complete_GCQRF} 
	Assume the index $\tau$ of interest lies in $[\tau_L, \tau_U]$ with bounded $\tau_L, \tau_U$. Denote $\theta_{s_n}(\tau) = \mathbb{P}(h_{s_n}(\omega, \tau))$ and $r_{1n}$ as defined in (\ref{eqn:ConvRate}). Assume $\limsup \mathbb{P} H_{s_n}(\omega)^2 < \infty$ where $H_{s_n}(\omega)$ is an envelope of $h_{s_n}(\omega, \tau)$. Suppose that Assumptions \ref{asmp:Randomness} - \ref{asmp:TreeBehavior_SubRate} hold. Then
	\begin{align}
		\frac{U_{n,s_n}h_{s_n}(\omega, \tau) - \theta_{s_n}(\tau)}{\sqrt{\frac{s_n^2r_{1n}^{2}}{n}}} \rsa \mathbb{G}_{1}, \label{eqn:G_1}
	\end{align}
	a mean-zero Gaussian process uniformly continuous with respect to $\rho(\tau, \tau')$ with covariance $\sigma_{1}(\tau, \tau')$.
\end{theorem}

The proof of Theorem \ref{thm:WeakCov_complete_GCQRF} can be found in the supplement material. Before establishing asymptotic results for randomized incomplete IDUPs, let us pause and compare the theorem above of randomized kernels to those in \cite{Heilig2001} for complete IDUPs of fixed kernels. It is noticeable that the limiting distribution of a randomized complete IDUP has the same form as that of some fixed kernel. To be more precise, the limiting distribution of a randomized complete IDUP is the same as that of a complete IDUP of some fixed kernel as long as the first order projected kernels of the randomized kernel and the fixed one are the same. This is as expected from the extended H-decomposition and the fact that we consider the \emph{complete} IDUP under the situation when the first order projection dominates. On the other hand, the additional randomness has a more substantial influence in the \emph{incomplete} case as shown in the next subsection.

\subsection{Asymptotics of Randomized Incomplete IDUPs}

In practice, the computational cost grows dramatically with $n$ and $s_n$, making it infeasible to grow trees on all $\n{s_n}$ possible subsamples. To ease this burden, consider growing trees on a subset of subsamples, leading to randomized \textit{incomplete} IDUP. As discussed in Sec \ref{sec:Prelim_UProcess}, we consider two different schemes for selecting subsamples \citep{Mentch2016, Chen2019, Peng2022}.

First consider bootstrap with replacement where the randomized incomplete IDUPs are defined by
\begin{align}
	U_{ n, s_n, m_n}h_{s_n}(\omega, \tau) = \frac{1}{m_n} \sum_{\beta \in (n, s_n)} M_{\beta} h_{s_n} (\mathcal{D}_{\beta}, \omega_{\beta};\tau) \label{eqn:RandomizedIncompleteIDUP_M_sec4.3}
\end{align}
where $M_{\beta}$, $\beta = 1, \dots, \n{s_n}$, is the number of times that the subsample $\mathcal{D}_{\beta}$, $\beta \in (n, s_n)$, is selected and $\bm{M}_{n, s_n} = \left(M_1, \dots, M_{\n{s_n}}\right ) \sim Multinomial\left (m_n; \frac{1}{\n{s_n}}, \dots, \frac{1}{\n{s_n}}\right )$. Then the approximation error of a randomized incomplete IDUP to a complete one is given by
\begin{align*}
	&  U_{ n, s_n, m_n}h_{s_n}(\omega, \tau) - U_{n,s_n} h_{s_n}(\omega, \tau) \\
	=& \frac{1}{\sqrt{m_n}} \ \frac{1}{\sqrt{\n{s_n}}}  \sum_{\beta=1}^{\n{s_n}} \sqrt{\frac{\n{s_n}}{m_n}}\left(M_{\beta} - \frac{m_n}{\n{s_n}}  \right) \left(h_{s_n}(\mathcal{D}_{\beta}, \omega_{\beta};\tau) - \theta_{s_n}(\tau)\right).
\end{align*}
The weak convergence of $U_{n, s_n, m_n}h_{s_n}(\omega, \tau)$ can be established by continuous mapping theorem when $U_{n,s_n}h_{s_n}(\omega, \tau) - \theta_{s_n}$ and $U_{ n, s_n, m_n}h_{s_n}(\omega, \tau) - U_{n,s_n} h_{s_n}(\omega, \tau)$ converges jointly with proper standardization.

As with the complete case, define 
\begin{align*}
	r_{hn}^{2} = \sup_{\tau} \zeta_{s_n, s_n}(\tau, \tau)
\end{align*}
and we assume that the variance of the tree kernel is of similar order across $\tau \in \Itau$ so that $h_{s_n}(\omega, \tau)$ can be standardized by $r_{hn}^{2}$.

\begin{assumption} \label{asmp:Var_Tree_SimilarMagnitude}
	There exists $d_{h}(\tau)$ s.t. $\zeta_{s_n, s_n}(\tau, \tau)/r_{hn}^{2}$ converges weakly to $d_{h}^{2}(\tau)$ for $\tau \in [\tau_L, \tau_U]$ and $d_{h}(\tau) \geq C_2 >0$  for some constant $C_2$.
\end{assumption}

Moreover, the behavior of the finite-dimensional marginals of the limiting process can be well-controlled by the following assumption.

\begin{assumption} \label{asmp:ControlFiniteDimMarginals}
	Suppose $\mathbb{P}\left[ \left|h_{s_n}(\omega, \tau) - \theta_{s_n}(\tau)\right|^{2k}\right]/\mathbb{P}^{2}\left[\left|h_{s_n}(\omega, \tau) - \theta_{s_n}(\tau) \right|^{k} \right]$ is uniformly bounded for $k = 2, 3$, all $s_n$ and all $\tau \in \Itau$.
\end{assumption}

Second, consider the randomized incomplete IDUP $U_{ n, s_n, N}h_{s_n}(\omega, \tau)$ defined in (\ref{eqn:RandomizedIncompleteIDUP_rho}) with respect to $\bm{\rho}_{n, s_n}$. Similarly, the approximation error of a randomized incomplete IDUP to a complete one is given by
\begin{align*}
	&	U_{ n, s_n, N}h_{s_n}(\omega, \tau) - U_{n,s_n} h_{s_n}(\omega, \tau)	\\
	=&	\frac{N}{\hat{N}} \left( 
		\frac{1}{\sqrt{N}} \ \frac{1}{\sqrt{\n{s_n}}}  \sum_{\beta=1}^{\n{s_n}} \sqrt{\frac{\n{s_n}}{N}}\left(\rho_{\beta} - \frac{N}{\n{s_n}}  \right) \left(h_{s_n}(\mathcal{D}_{\beta}, \omega_{\beta};\tau) - \theta_{s_n}(\tau)\right)
	 \right)
\end{align*}
and the weak convergence of $U_{ n, s_n, N}h_{s_n}(\omega, \tau)$ can be established in the same way under the same set of assumptions above.

Define
\begin{align}
	\sigma_{h}^{\omega}(\tau, \tau') = d_{h}(\tau)d_{h}(\tau')\lim_{n} \frac{\zeta_{s_n, s_n}(\tau, \tau')}{\sqrt{\zeta_{s_n, s_n}(\tau, \tau) \zeta_{s_n, s_n}(\tau', \tau')}} . \label{eqn:Cor_TreeKernel}
\end{align}
and let $\mathbb{G}_{h}$ denote a mean-zero Gaussian process with covariance $\sigma_{h}^{\omega}(\tau, \tau')$.

\begin{theorem}[Weak Convergence of Randomized Incomplete IDUPs]\label{thm:WeakCov_IDUP_incomplete} 
	Assume the index $\tau \in [\tau_L, \tau_U]$ with bounded $\tau_L, \tau_U$. Denote $\theta_{s_n}(\tau) = \mathbb{P}(h_{s_n}(\omega, \tau))$. Assume $\limsup \mathbb{P} H_{s_n}(\omega)^2 < \infty$ where $H_{s_n}(\omega)$ is an envelope of $h_{s_n}(\omega, \tau)$. 	
	Suppose that Assumptions \ref{asmp:Randomness} - \ref{asmp:ControlFiniteDimMarginals}  hold.
	\begin{enumerate}
		\item Suppose $U_{ n, s_n, m_n}h_{s_n}(\bmo, \tau)$ is an randomized incomplete IDUP defined by (\ref{eqn:RandomizedIncompleteIDUP_M}). Let $\alpha	= \lim_{n} \left(s_n^2r_{1n}^{2}/n\right)/\left(s_n^2r_{1n}^{2}/n + r_{hn}^{2}/m_n\right).$
		Then
		\begin{align*}
			\frac{U_{n, s_n, m_n}h_{s_n}(\bmo, \tau) - \theta_{s_n}(\tau)}{\sqrt{\frac{s_n^2r_{1n}^{2}}{n} + \frac{r_{hn}^{2}}{m_n}}} \rsa \sqrt{\alpha}\mathbb{G}_{1} + \sqrt{1 -\alpha} \mathbb{G}_{h},
		\end{align*}
		a mean-zero Guassian process with covariance given by $\alpha \sigma_{1}(\tau, \tau') + (1-\alpha) \sigma_{h}^{\omega}(\tau, \tau')$ where $\mathbb{G}_{h}$ and $\sigma_{h}^{\omega}(\tau, \tau')$ are defined above and $\sigma_{1}(\tau, \tau')$ and $\mathbb{G}_{1}$ in (\ref{eqn:Cor_1stOrder}) and (\ref{eqn:G_1}) respectively.
		
		\item Suppose $U_{n, s_n, N}h_{s_n}(\bmo, \tau)$ is an randomized incomplete IDUP defined by (\ref{eqn:RandomizedIncompleteIDUP_rho}).  Then the same results as in the first part hold by replacing $m_n$ with $N$.
	\end{enumerate}
\end{theorem}

The proof of Theorem \ref{thm:WeakCov_IDUP_incomplete} is deferred to the supplement material. Along the proof, we have the following corollary on the independence of the limiting processes of the randomized complete IDUP and the approximation error of the randomized incomplete IDUP to the complete one when the number of selected subsamples is much smaller than the total number of possible subsamples.

\begin{corollary} \label{cor:IndependentLimitingProcesses}
	Suppose conditions in Theorem \ref{thm:WeakCov_IDUP_incomplete} are met. When $m_n/\n{s_n} \rightarrow 0$ or $N/\n{s_n} \rightarrow 0$, the limiting processes $\mathbb{G}_{1}$ and $\mathbb{G}_{h}$ are independent.
\end{corollary}

The effect of the additional randomness is more clearly unraveled in Theorem \ref{thm:WeakCov_IDUP_incomplete}. When the number of selected subsamples is large in the sense that $m_n/\n{s_n}$ or $N/\n{s_n} \rightarrow c \in (0,1]$, then the rate of convergence is dominated by $\sqrt{s_n^2r_{1n}^2/n}$ and $\alpha = 1$. In another word, the randomized incomplete IDUP is close to a complete one and the limiting process is determined by that of the randomized complete IDUP. In practice, it will be more realistic to have $m_n/\n{s_n}$ or $N/\n{s_n} \rightarrow 0$. In such circumstance, to quantify the variation of the prediction, we should take into account both the covariance $\sigma_{1}$ of the first order projected kernel and the covariance $\sigma_{h}^{\omega}$ of the randomized tree kernel where the expectation is not only over the training data $\mathbb{D}_{n} = \{\bm{Z}_1, \dots, \bm{Z}_n\}$ but also over the additional randomness $\omega$. In the extreme scenario when $N$ or $m_n$ is small such that $\alpha = 0$, then the approximation error of a randomized incomplete IDUP to the complete counterpart dominates and the covariance is solely determined by $\sigma_{h}^{\omega}$.

\section{Feature Importance}\label{sec:vimp}

While black-box machine learning tools, such as random forests, excel in prediction accuracy, they encounter challenges in interpretation, a crucial aspect for decision making and resource allocation. Recently, researchers have turned to out-of-sample prediction performance as a surrogate measure to assess the conditional importance of individual features or groups of features in relation to the target response, given the rest features. 

In a general methodology, two models are established: one incorporating all features and another with the effects of the features under investigation muted and the rest the same as in the first model. Feature importance is then ranked based on change in out-of-sample prediction performance. As elucidated by \cite{Hooker2021}, the impact of features of interest can be muted through techniques such as removal, permutation, or resampling from the estimated conditional distribution given the remaining features. In contrast to removal or permutation, the estimating-and-resampling approach is more resilient against erroneously identifying noise features as significant, but it can be time-consuming, particularly with a large feature dimension. An alternative approximation involves replacing features of interest with knockoff variables \citep{Candes2018}, which remain independent of the response given the original features while preserving between-feature relationships

It is essential to note that certain approaches, like the permutation test proposed by Breiman \citep{Breiman2001}, involve constructing \textit{only one model} with all features and, subsequently, evaluating the performance of \textit{this} model on a dataset where features of interest are permuted unconditionally, rather than evaluating the performance of a new model \textit{refitted} to the permuted dataset. Approaches like this are more appropriate for assessing the feature importance to the constructed, fixed model rather than the relationship between the features and the response. When the latter is the focus, despite of the low computational cost, the change of predictive performance measured in this way is subject to the ability to extrapolate of the model under consideration, leading to potentially biased interpretation.
Here, we advocate for the reconstruction approach when the target is not at a fixed model. More detailed discussions and other alternative remedies can be found in \cite{Hooker2021}.

Moreover, both models should be tuned as carefully as possible so as to be considered as the models optimizing the predictive performance \citep{Williamson2021}. Otherwise, noise features can be falsely identified as important due to their implicit regularization effect \citep{Mentch2022}. While a hold-out test set or cross validation is commonly exploited for tuning, random forests come with a ready-to-use metric called out-of-bag (OOB) error. Specifically, with subsampling or bootstrapping, only a proportion of training observations are used for each tree and those left out are called OOB observations. For each observation $\bm{x}$, the OOB prediction at $\bm{x}$ is obtained by averaging the predictions of trees where $\bm{x}$ is considered as OOB. The corresponding error rate is then referred to as the OOB error. This can be calculated during training without incurring significant additional computational cost and performs similarly to using a test set of the same size as the training one \citep{Breiman2001, ESL}.

Considering these, we propose to measure feature importance as follows.
First grow a tuned GCQRF with all features available and obtain the OOB I-QLoss denoted by $\widehat{L}_{\rho, \text{full}}$. Next, for a feature of interest, grow another separately tuned GCQRF with this feature dropped, permuted or replaced by its knockoff counterpart and calculate its OOB I-QLoss $\widehat{L}_{\rho, \text{modified}}$. 
The larger $\widehat{L}_{\rho, \text{modified}} - \widehat{L}_{\rho, \text{full}}$ is, the more important this feature is conditional on the rest. When the dimension of features under investigation is large, it is more efficient to group them based on domain knowledge.
Alternatively, crossing fitting \citep{Williamson2021} can be utilized to stabilize the feature importance measure as illustrated in Algorithm \ref{alg:VIMP_Cross}.

\begin{algorithm}[!tb]
	\caption{Conditional Feature Importance Ranking with Cross Fitting}
	\label{alg:VIMP_Cross}
	\begin{algorithmic}[1]
	\Require Training data $\mathcal{D}_{n}$, number $K$ of folds for cross fitting, features $\mathcal{J}$ of interest, quantile levels $\tau$, tuning space
	\State Divide $\mathcal{D}_{n}$ into $K$ folds
	\For{$k = 1, \dots, K$}
		\State On the rest $K-1$ folds of data other than the $k$-th one, grow a GCQRF tuned with respect to OOB I-QLoss and measure the its I-QLoss $\widehat{L}_{\rho, 0}^{k}$ on the $k$-th fold leftout.
		\State On the rest $K-1$ folds of data other than the $k$-th one, grow another separately tuned GCQRF with features $\mathcal{J}$ dropped, permuted or resampled from knockoff counterparts, and measure the its I-QLoss $\widehat{L}_{\rho, \mathcal{J}}^{k}$ on the $k$-th fold leftout.
	\EndFor
	\State Obtain the conditional feature importance ranking according to the mean cross-fitted I-QLoss
	\begin{align*}
		\frac{1}{K}\sum_{k=1}^{K} (\widehat{L}_{\rho, \mathcal{J}}^{k} - \widehat{L}_{\rho, 0}^{k}).
	\end{align*}

	\end{algorithmic}
\end{algorithm}

When the target is to assess feature importance for predicting the conditional mean with performance measured by squared error, \cite{Hooker2021} showed that the three approaches above for muting effects of features targeted at the same quantity. As shown in the Proportion \ref{prop:vimp}, this is also true for feature importance for predicting conditional quantiles in case of no censoring when performance can be measured based on the standard I-QLoss. 

\begin{proposition} \label{prop:vimp}
	Suppose $\bm{X} = (\bm{X}_1, \bm{X}_2)$. Let $\bm{X}^{*} = (\bm{X}_1^{*}, \bm{X}_2)$ where $\bm{X}_1^{*} \perp T |\bm{X}_2$. For each $\tau$, 
	\begin{align*}
		\argmin_{m}\mathbb{P}_{Y|\bm{X}^{*}} \rho_{\tau}(T - m) = Q_{T|\bm{X}_2}(\tau)
	\end{align*}		
	where $Q_{T|\bm{X}_2}(\tau)$ is the $\tau$-th conditional quantile of $T$ given $\bm{X}_2$. When $\bm{X}_1$ is removed, permuted or resampled from the conditional distribution $\bm{X}_1 |\bm{X}_2$, the conditional feature importance of $\bm{X}_1$ given $\bm{X}_2$ is then given by
	\begin{align*}
		&	\mathbb{P}\left[\int_{\tau} \tau(1-\tau)  \left[ \rho_{\tau}(T - Q_{T|\bm{X}_2}(\tau)) -  \rho_{\tau}(T - Q_{T|\bm{X}}(\tau))\right]  d\tau 	\right].
	\end{align*}
\end{proposition}

\section{Numerical Analysis}\label{sec:num}

In this section, we carry out numerical analysis in \texttt{R} to compare the predictive accuracy of GCQRF to existing alternatives and illustrate the proposed conditional feature importance measure on both simulated and real data. 

\subsection{Simulated Data}
\subsubsection{Predictive Accuracy}\label{sec:num_sim_predictive}


We begin by describing the data generating process for comparing predictive performance. Consider the true response given by
\begin{align*}
	T = f(\bm{X}) + g(Z)  + \epsilon
\end{align*}
where $f(\cdot), \ g(\cdot)\in \mathbb{R}$, $\bm{X} = (X_1, X_2, \dots, X_p)' \in \mathcal{X} \subset \mathbb{R}^p$, $Z \sim Bernoulli(0.5)$, $\epsilon \sim N(0, \sigma^2)$ and $\bm{X}$, $Z$ 
and $\epsilon$ are independent. 
Define the signal-to-noise ratio (SNR) of this model by
\begin{align*}
	\SNR = \frac{\Var(f(\bm{X})) + \Var(g(Z))}{\sigma^2}.
\end{align*}
The larger the SNR, the higher the quality of the data in the sense that the more variation of the response can be explained by the available features. 
Given a pre-specified value of SNR, the noise level $\sigma^2$ is determined by
\begin{align*}
	\sigma^2 = \frac{\Var(f(\bm{X})) + \Var(g(Z))}{\SNR}.
\end{align*}
We consider 10 values of SNRs ranging from 0.05 to 6, equally spaced on the $\log$ scale. This allows us to exam the performance of the proposed method over data of different qualities. 


For each SNR, we analyze data where the true response is generated either from linear models or nonlinear models by specifying the forms of 
$f(\bm{X})$.  More specifically, we consider 3 different forms of $f(\bm{X})$ as follows.  
\begin{itemize}
	\item \textbf{Linear}: The first one is a linear model studied in \cite{Hastie2020}. In particular, $f(\bm{X}) = \bm{X\beta}$ where $\bm{X} \sim MN(\bm{0}, \Sigma)$, $\Sigma_{ij} = \rho^{|i-j|}$ with $\rho = 0.35$ and the first $r=5$ terms in $\bm{\beta}$ equal to 1 and the rest equal to 0, corresponding to the \texttt{beta.type} = 2 setting in \cite{Hastie2020}.
	
	\item \textbf{Nonlinear}: Next, consider a nonlinear model studied in \cite{Scornet2017} given by
	\begin{align*}
		f(\bm{X}) =& X_1 + 4(X_2 -0.5)^2 + \frac{\sin(2\pi X_3)}{2 - \sin(2\pi X_3)} + \\
				   & \sin(2\pi X_4) + 2\cos(2\pi X_4) + 3\sin^2(2\pi X_4) + 4\cos^2(2\pi X_4)
	\end{align*}
	where $\bm{X} \sim Unif([0,1]^p)$, $p\geq 4$, with first $r = 4$ signal features. 
\end{itemize}


Additionally, we employ two versions of $g(Z)$ to differentiate whether the features involve only location-shift effects, such as in the accelerated failure time (AFT) model, or not.
\begin{itemize}
	\item Homogeneous setting (HOMO): $g(Z) = Z$. In this case, the variance of $T$ is constant and the true conditional quantiles of $T$ are given by
\begin{align*}
	Q(\tau|\bm{X}, Z) = f(\bm{X}) + Z + Q_{\epsilon}(\tau)
\end{align*}
where $Q_{\epsilon}(\tau)$ is the $\tau$-th quantile of $\epsilon \sim N(0, \sigma^2)$ and all features entail location-shift effects.
	\item Heterogeneous setting (HETE): $g(Z) = \xi Z$ where $\xi \sim N(0,1)$,  independent of $\bm{X}, \ Z, \ \epsilon$. Then the variance of $T$ depends on the $Z$ and the true conditional quantiles of $T$ are given by
	\begin{align*}
		Q(\tau|\bm{X}, Z) = f(\bm{X})  + Q_{\epsilon + \xi I(Z=1) }(\tau)
	\end{align*}
	where $Q_{\epsilon + \xi I(Z=1) }(\tau)$ is the $\tau$-th quantile of $\epsilon + \xi I(Z=1) \sim N(0, \sigma^2 + I(Z=1))$. 
\end{itemize}

We consider censoring variable $C$ whose distribution depends on $Z$: $C\sim Unif(c_{l1}, c_{u1})$ if $Z = 1$ and $C \sim Unif(c_{l0}, c_{u0})$ otherwise. Here $c_{l1}, c_{u1}, c_{l0}, c_{u0}$ are simulated such that the resulting censoring rate is within 1\% of a targeted censoring rate. In the following, we consider censoring rate = 30\%.

The training data is in the form of $\mathcal{D}_{n} = \{(\bm{X}_i, Z_i, \delta_i, Y_i):  Y_i = \min(T_i, C_i), \ \delta_i = I(T_i \leq C_i), \ i = 1,\dots, n \}$ with the following 2 problem settings considered.
\begin{itemize}
	\item Low: $n = 100$, $p = 10$
	\item High: $n = 100$, $p = 200$
\end{itemize}
Recall that $p$ is the dimension of $\bm{X}$. Together with $Z$, there are in total $p + 1$ features in each setting. All in all, for each SNR, we consider 8 settings as summarized in Table \ref{tbl:sim_settings}.

\begin{table}[ht]
	\caption{Simulation settings under consideration for each SNR}
	\label{tbl:sim_settings}
	\centering
	\begin{tabular}{|c|c|c|}
	\hline
	$f(\bm{X})$	&	$g(Z)$	&	data dimensions	\\
	\hline
	Linear, Nonlinear	&	HOMO, HETE	&	Low, High	\\
	\hline
	\end{tabular}
\end{table}


We compare the performance of proposed GCQRF to the following alternative methods.  The first two methods provide estimates of the cumulative hazard function, and quantiles are estimated by inverting these estimates. 
\begin{itemize}
	\item Nelson-Aalen (N-A) estimator: N-A estimator does not consider any features available and is used as the baseline for comparison.
	\item Random survival forest (rsf) \citep{Ishwaran2008}: implemented in the package \texttt{randomForestSRC} with log rank test statistic used as the splitting criterion
	\item Peng-Huang (PH) model \citep{Peng2008}: a censored quantile regression model with linear assumptions implemented in \texttt{R} package \texttt{quantreg}.
	\item PH-oracle: the same as PH except that only the true signal features are included.
	\item Generalized random forest (grf) \citep{Athey2019}: the quantile forest implemented in the package \texttt{grf}. Internal nodes are split using a multiclass classification rule based on gradient-based relabeled responses. More details can be found in Section 5 of \cite{Athey2019}. With the observed response $y$, grf does not adjust for censoring.
	\item Quantile regression forest (qrf) \citep{Meinshausen2006}: compared to grf, the CART criterion is used as the splitting rule. This is also implemented in the package \texttt{grf} with no adjustment for censoring.
	\item grf-oracle and qrf-oracle: the respective oracle versions of grf and qrf where the true response $t$ is supplied. This is not applicable in practice and only serves as a reference for comparison.
	\item Censored quantile regression forest (crf) \citep{Li2020}: based on a forest ignoring censoring, crf estimates conditional quantiles by employing post-forest-construction adjustment for censoring. We implement this using codes in the package \texttt{cERF} and call the method crf-qrf or crf-grf depending on the forest used. Notice that crf is a local method for one specific $\tau$ while our target is to exam the overall predictive accuracy over an interval of $\tau$, so a sequence of forests are constructed over a grid of $\tau$.
\end{itemize}
In addition, we also include the true conditional quantiles (TrueQ) from the data generating process as a reference.

Performance is evaluated on an independent test set of the same size as the training one.
With known true conditional quantiles in the simulations, we can compare model performance based on the mean squared error on conditional quantiles (QMSE). Define QMSE for a given $\tau$ by  
\begin{align*}
	\widehat{L}_{2}(\widehat{Q}, \tau) = \frac{1}{n}\sum_{i=1}^{n} \tau(1 - \tau) \left (Q(\tau|\bm{X}_i) - \widehat{Q}(\tau|\bm{X}_i)\right )^2.
\end{align*}
To summarize the performance over an interval of $\tau$, we consider the integrated QMSE (I-QMSE)
\begin{align*}
	\widehat{L}_{2}(\widehat{Q}) = (\tau_U -\tau_L)^{-1}\int_{\tau_L}^{\tau_{U}} \widehat{L}_2(\widehat{Q}, \tau)d\tau
\end{align*}
which is approximated over a grid $S_{\tau} = \{ 0 = \tau_0 < \tau_1 = \tau_L < \dots < \tau_{n_\tau} = \tau_{U} < 1 \}$. Here we consider $\tau_L = 0.05$ and $\tau_U = 0.5$ with increment 0.05.

When true quantiles are unknown, we can use I-QLoss $\widehat{L}_{\rho}(\widehat{Q})$ 
\begin{align*}
	\widehat{L}_{\rho}(\widehat{Q}) = 
		(\tau_U - \tau_L)^{-1} \int_{\tau_L}^{\tau_U} 
			\frac{1}{n} \sum_{i=1}^{n}  \frac{\Delta_i}{\widehat{G}(Y^u_i  |\bm{X}_i)} \tau(1-\tau) \rho_{\tau}\left (Y_i^u - \widehat{Q}(\tau |\bm{X}_i)\right ) 
		d\tau
\end{align*}
discussed in Section \ref{sec:Model} to evaluate the performance of methods other than grf-oracle and qrf-oracle where, with a prespecified $u$ less than the upper bound of $C$, $Y^u_i = T^u_i \wedge C_i = Y_i \wedge u$, $\Delta_i = I(T^u_i \leq C_i) = I(Y_i >u) + I(Y_i \leq u, \delta_i = 1)$ and $\widehat{G}(t|\bm{X})$ is the estimation of the conditional survival function of censoring $Pr(C>t|\bm{X})$ from a rsf tuned with respect to C-index. For grf-oracle and qrf-oracle where the true response $t$ is available, the performance can be evaluated by the standard I-QLoss
\begin{align*}
	\widehat{L}_{\rho}(\widehat{Q}) = 
		(\tau_U - \tau_L)^{-1} \int_{\tau_L}^{\tau_U} 
			\frac{1}{n} \sum_{i=1}^{n}  \tau(1-\tau) \rho_{\tau}(T - \widehat{Q}(\tau |\bm{X}) 
		d\tau
\end{align*}
which is the I-QLoss with respect to the standard quantile loss $\rho_{\tau}(v) = v(\tau - I(v < 0))$. In another word, grf-oracle and qrf-oracle are oracle in the sense that no censoring adjustment is needed for either prediction or performance evaluation.
For notation brevity and focusing on the methods under comparison, we have only included the dependence of loss functions $\hat{L}_{2}(\widehat{Q})$ and $\hat{L}_{\rho}(\widehat{Q})$ on $\widehat{Q}$ and omitted their dependence on other quantities such as the true conditional quantiles $Q(\tau |\bm{X})$ and the quantile levels of interest $[\tau_L, \tau_U]$.

For each method $\widehat{Q}$ under investigation, we define the relative I-QMSE by
\begin{align*}
	\text{relative I-QMSE} = \frac{\widehat{L}_{2}(\widehat{Q})}{\widehat{L}_{2}(\widehat{Q}_{\NA})}
\end{align*} 
and relative I-QLoss by
\begin{align*}
	\text{relative I-QLoss} = \frac{\widehat{L}_{\rho}(\widehat{Q})}{\widehat{L}_{\rho}(\widehat{Q}_{\NA})}
\end{align*} 
where the denominators are the I-QMSE and I-QLoss of the Nelson-Aalen estimator respectively.

Recent studies provide evidence that tuning is important for random forests \citep{Scornet2017, Probst2019, Mentch2020, Zhou2023}. Thus, we tune all forest-based methods with respect to OOB I-QLoss over the following tuning space.
\begin{itemize}
	\item \texttt{mtry} as a proportion of $p$: 0.1, 0.3, 0.5, 0.7, 0.9.
	\item Subsampling rate $\gamma$ as a proportion of the training size: 0.3, 0.5, 0.7, 0.9. 
	\item \texttt{nodesize} = $n^\eta$ where $\eta$ takes 5 values equally spaced between $\frac{\log(5)}{\log(n)}$ and $\frac{\log(n/4)}{\log(n)}$. 
\end{itemize}
For models implemented using the \texttt{grf} package, we also tune the parameter \texttt{honesty} over \texttt{TRUE} or \texttt{FALSE}. When \texttt{honesty} is \texttt{TRUE}, a tree is grown on part of the selected subsample with rest used for prediction. Otherwise, the entire subsample is used for both purposes. In our simulation, an equal-size splitting is considered when \texttt{honesty} is \texttt{TRUE}. To reduce the cost of tuning, all forests are tuned with the number of trees equal to 100. Once the parameters above are tuned, each forest is grown with the number of trees set to the default value, namely, 500 for rsf and GCQRF and 2000 for the rest whose implementations depend on the \texttt{grf} package.

\begin{figure}[ht!]
	\centering
    \includegraphics[width = 0.99\textwidth]{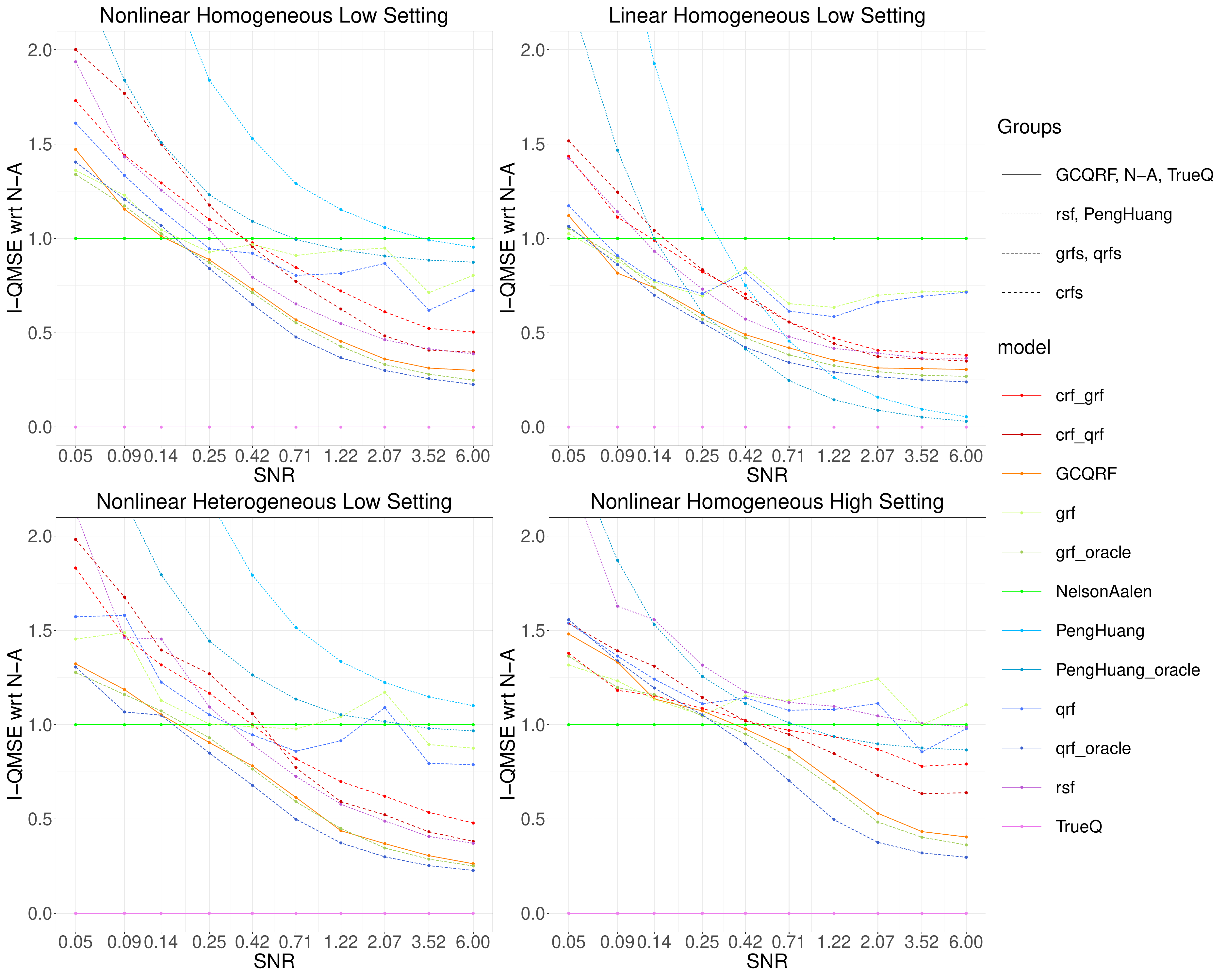}
	\caption{Comparison on predictive performance based on \textbf{test relative I-QMSE} in the settings of Nonlinear Homogeneous Low Dimension (upper left), Linear Homogeneous Low Dimension (upper right), (c) Nonlinear Heterogeneous Low Dimension (lower left) and Nonlinear Heterogeneous High Dimension (lower right). Results for the rest settings can be found in the supplement.}
	\label{fig:pfm_main_IQMSE}
\end{figure}

Figure \ref{fig:pfm_main_IQMSE} shows the average test relative I-QMSE versus SNR for several settings. The full results with $\pm 1$ error bars are deferred to the supplement material. 
The lower the relative I-QMSE, the better the performance.
Overall, the proposed GCQRF is promising and competitive for a wide range of settings.

Results for the Nonlinear HOMO Low setting are shown in the upper left figure in Figure \ref{fig:pfm_main_IQMSE}. At low SNRs from 0.05 to 0.14, the test relative I-QMSEs of all methods are above 1, implying that N-A estimator which does not consider any features available outperforms all other methods under consideration.
For SNRs between 0.25 and 6, the proposed GCQRF exhibits promising and competitive performance with the test relative I-QMSE of GCQRF less than those of the rest methods under consideration except grf-oracle and qrf-oracle which make use of the true underlying response and are not applicable in practice.
When censoring is ignored, grf and qrf have the highest test relative I-QMSE and thus perform the worst among all forest-based methods. 
Comparing crf-qrf and crf-grf to qrf and grf respectively, the post-forest-growing censoring adjustment in crf lowers the error of the corresponding forest ignoring censoring. However, this adjustment is not as effective as GCQRF where censoring is adjusted while the forest is being grown. In addition, crf targets at estimating a particular quantile and thus has to be grown multiple times when the interest is to estimate the quantile process, while only one forest is necessary for GCQRF. 
rsf is outperformed by GCQRF at all SNRs above 0.14. With the log rank statistic used for splitting criterion, rsf searches for the best split to distinguish the survival distributions of the two resulting nodes but may not be sensitive enough to changes in quantiles. 

In comparing results in this Nonlinear HOMO Low setting to the Linear HOMO Low setting, forest-based approaches are more robust to the underlying data generating process compared with the linear methods PH and PH-oracle. In the Nonlinear scenario, PH and PH-oracle yield higher test relative I-QMSEs compared to forest-based methods. On the other hand, when the linear assumption is satisfied, PH and PH-oracle improve dramatically as the SNR increases, yielding lower test relative I-QMSEs than forest-based approaches at medium to high SNRs. Among the forests other than grf-oracle and qrf-oracle, GCQRF remains to be the best in this Linear scenario.

Comparing the upper left and lower left figures in Figure \ref{fig:pfm_main_IQMSE}, results for the Nonlinear HETE Low setting are similar to those for the Nonlinear HOMO Low setting discussed above. In particular, the proposed GCQRF has lower test relative I-QMSE than other methods except grf-oracle and qrf-oracle.

Lower right figure in Figure \ref{fig:pfm_main_IQMSE} shows the result for the Nonlinear HOMO High setting. With the additional high dimensional noise features, test relative I-QMSEs of all methods increase to different extensions and stronger signals are required to outperform the N-A estimator while GCQRF, grf-oracle and qrf-oracle are more robust with smaller increase in the test I-QMSE compared with rsf and crfs. Other than grf-oracle and qrf-oracle, GCQRF has lower test relative I-QMSE and outperforms the rest methods for SNRs larger than 0.25.

\begin{figure}[ht!]
	\centering
    \includegraphics[width = 0.99\textwidth]{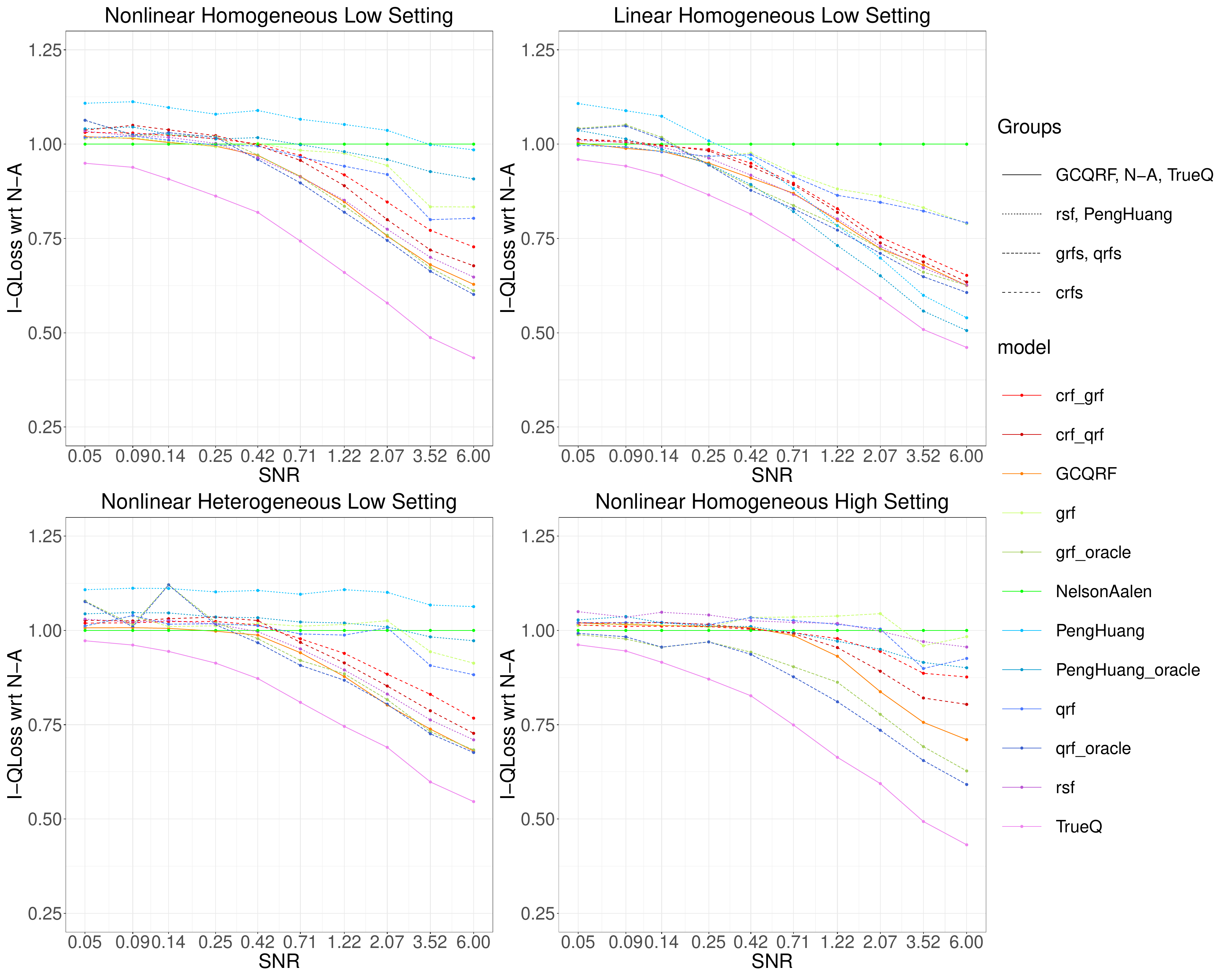}
	\caption{Comparison on predictive performance based on \textbf{test relative I-QLoss} in the settings of Nonlinear Homogeneous Low Dimension (upper left), Linear Homogeneous Low Dimension (upper right), (c) Nonlinear Heterogeneous Low Dimension (lower left) and Nonlinear Heterogeneous High Dimension (lower right). Results for the rest settings can be found in the supplement.}
	\label{fig:pfm_main_IQLoss}
\end{figure}

Figure \ref{fig:pfm_main_IQLoss} shows the results for the same four settings above but with respect to test relative I-QLoss. Results are similar to those with I-QMSE. When signals are strong enough for GCQRF to outperform N-A estimator, GCQRF performs similar to or better than other forest-based approaches except for the oracle ones. PH and PH-oracle perform best when linear assumptions are satisfied and the SNR is large enough.

\subsubsection{Case Study on Feature Importance}\label{sec:num_sim_vimp}

Next, we carry out simulations to illustrate the proposed conditional feature importance measure. Consider the following two models similar to the linear model studied in Section \ref{sec:num_sim_predictive}.
\begin{itemize}
	\item Homogeneous (HOMO) error
	\begin{align}
		T = \bm{X}_1^{T}\bm{\beta}_1 + \bm{X}_2^{T}\bm{\beta}_2 + \gamma Z + \sigma \epsilon \label{eqn:vimp_homo}
	\end{align}
	
	\item Heterogeneous (HETE) error
	\begin{align}
		T = \bm{X}_1^{T}\bm{\beta}_1 + \bm{X}_2^{T}\bm{\beta}_2 + \gamma \xi Z + \sigma \epsilon \label{eqn:vimp_hete}
	\end{align}
\end{itemize}
where
\begin{align*}
\bm{X}=	\begin{pmatrix}
	\bm{X}_1 \\ \bm{X}_2
	\end{pmatrix}
	&\sim N\left(
		\begin{pmatrix}
			\bm{0} \\		\bm{0}
		\end{pmatrix}
		,
		\Sigma = 
		\begin{pmatrix}
			\Sigma_{11}	&	\Sigma_{12}	\\
			\Sigma_{21}	&	\Sigma_{22}
		\end{pmatrix}
	\right), \\
	Z &\sim Bernoulli(p = 0.5), \quad \epsilon \sim N(0,1), \quad \xi \sim N(0, 1),
\end{align*}
and $\bm{X}$, $Z$, $\epsilon$ and $\xi$ are independent. Same as in Section \ref{sec:num}, (i) the censoring $C$ depends on $Z$: $C \sim Unif(c_{l1}, c_{u1})$ if $Z = 1$ and $C \sim Unif(c_{l0}, c_{u0})$ otherwise where $c_{l1}$, $c_{u1}$, $c_{l0}$, $c_{u0}$ are simulated such that the censoring rate is around 30\%, and (ii) the data is in the form of $\mathcal{D}_{n} = \{(\bm{X}_i, Z_i, \delta_i, Y_i):  Y_i = \min(T_i, C_i), \ \delta_i = I(T_i \leq C_i), \ i = 1,\dots, n \}$.

For these simple models, we can derive a simplified expression for the true conditional quantiles and thus the targeted conditional feature importance. Though not applicable in practice, we can also measure the conditional feature importance by change in I-QMSE in these simulations.

Generally speaking, $\bm{X}_1$ and $\bm{X}_2$ entail location-shift effect for both (\ref{eqn:vimp_homo}) of HOMO error and (\ref{eqn:vimp_hete}) of HETE error. The conditional importance of $\bm{X}_1$ given $\bm{X}_2$ and $Z$ depends on the magnitudes of $\bm{\beta}_1$ and the conditional variance of $\bm{X}_1$ given $\bm{X}_2$. Thus, for the same $\bm{\beta}_1$, the conditional importance of $\bm{X}_1$ would be smaller in the case when $\bm{X}_1$ is correlated with $\bm{X}_2$ than in the case when $\bm{X}_1$ is independent of $\bm{X}_2$. Intuitively, the effect of $\bm{X}_1$ is partially captured by $\bm{X}_2$ when they are correlated.

The two models (\ref{eqn:vimp_homo}) and (\ref{eqn:vimp_hete}) mainly differ in the effect of $Z$. In the HOMO case, $Z$ also has a location-shift effect, which is closely related to a mixture of normal distributions that are the same in spread but different in center. In contrast, $Z$ shifts the scaling in the HETE case, and its effect on the conditional quantiles is tied to another mixture of normal distributions which, instead, are the same in center but different in spread.

To be more concrete, take
\begin{align*}
	\bm{\beta}' = (\bm{\beta}'_1, \bm{\beta}'_2) = (1.4, 1.4, 1.4, 0, 1, 2), \quad \gamma = 2, \quad \sigma = 0.1
\end{align*}
and $\Sigma = [\sigma_{ij}]$ with $\sigma_{ii} =1$, for $i=1, \dots, 6$, $\sigma_{ij} = 0$ for $i \neq j$ except that $\sigma_{12} = \sigma_{21} = \rho$. We consider $\rho = 0$ for the independent case and $\rho = 0.9$ for the correlated case. In addition, we consider $\tau \in [0.1, 0.3]$ or $\tau \in [0.4, 0.6]$.

\begin{figure}[!ht]
	
	\centering
	\begin{subfigure}[!ht]{0.45\textwidth}
         \centering
         \includegraphics[width = \textwidth]{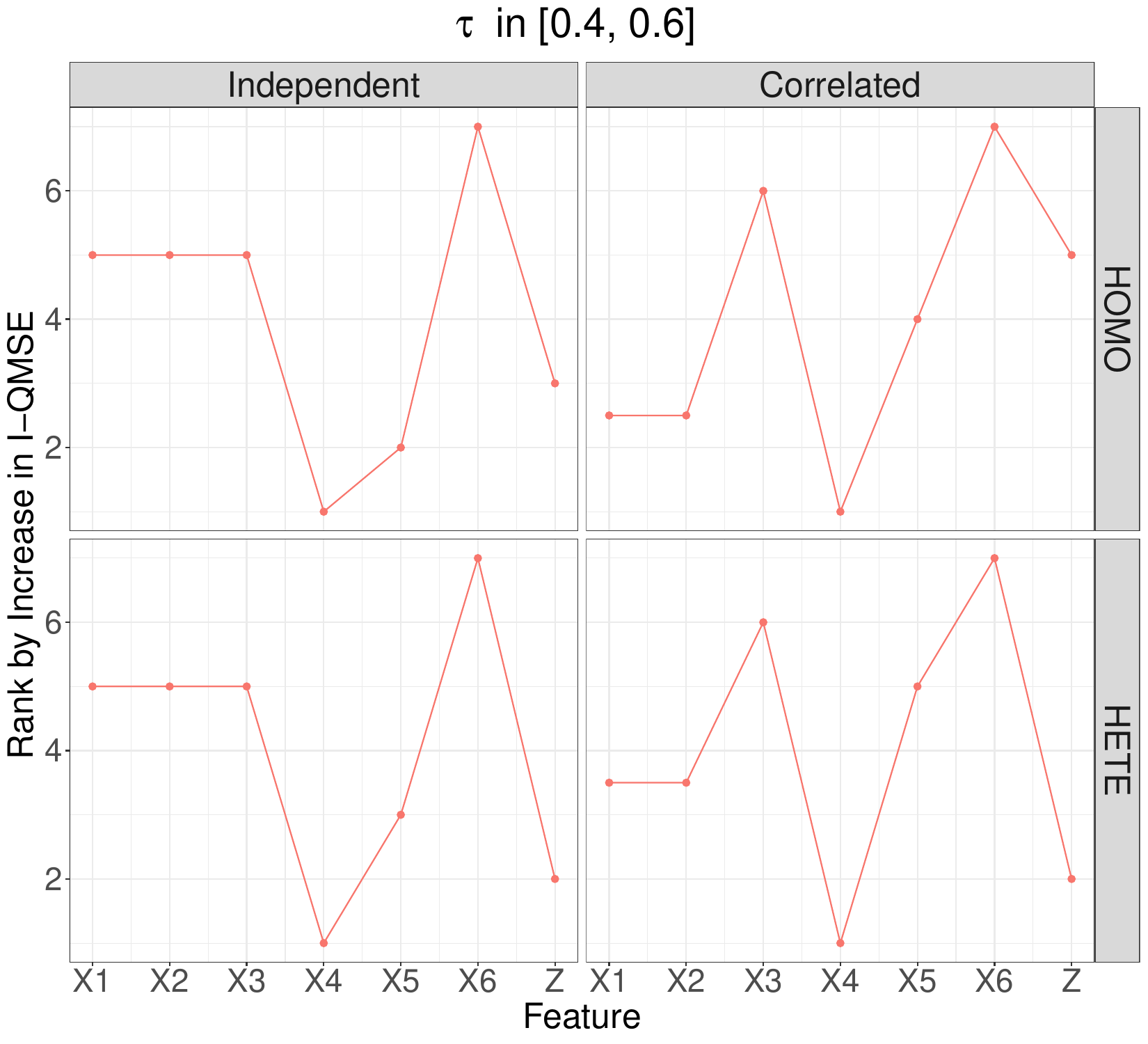}
         \caption{$\tau \in [0.4, 0.6]$}
         \label{fig:vimp_rank_true_Tau40To60}
     \end{subfigure}
     \begin{subfigure}[!ht]{0.45\textwidth}
         \centering
         \includegraphics[width = \textwidth]{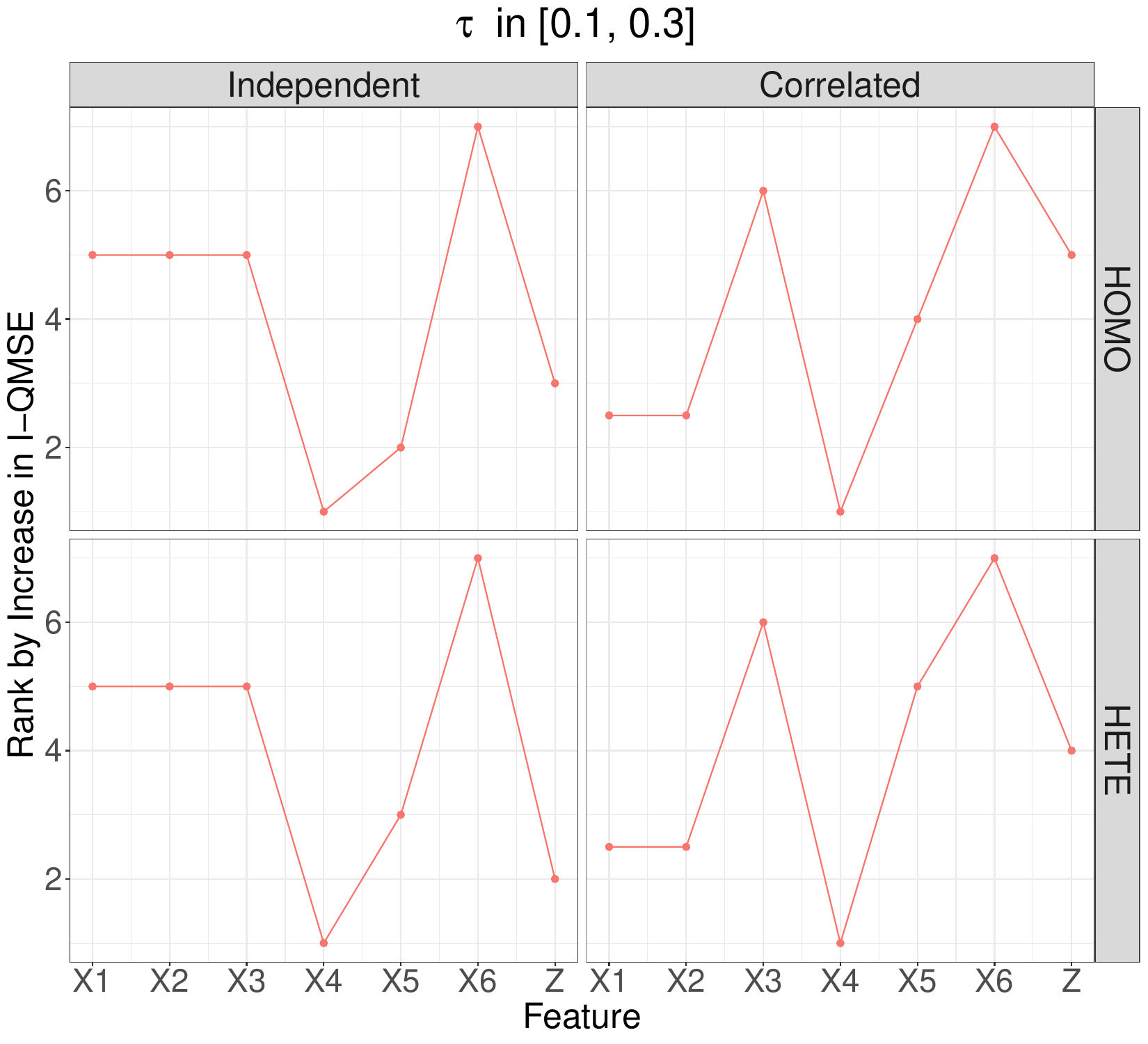}
         \caption{$\tau \in [0.1, 0.3]$}
         \label{fig:vimp_rank_true_Tau10To30}
     \end{subfigure}

	\caption{\textit{True} conditional feature importance ranking based on increase in I-QMSE with (a) $\tau \in [0.4, 0.6]$ and (b) $\tau \in [0.1, 0.3]$. In both (a) and (b), the left and right columns correspond to the independent case ($\rho = 0$) and correlated case ($\rho = 0.9$) respectively with data generated from the model with homogeneous error (top row) or heterogeneous error (bottom row). }
	\label{fig:vimp_rank_true}
\end{figure}

With these values, the true conditional feature importance ranking based on increase in I-QMSE for each feature is plotted in Figure \ref{fig:vimp_rank_true_Tau40To60} for $\tau \in [0.4, 0.6]$ and in Figure \ref{fig:vimp_rank_true_Tau10To30} for $\tau \in [0.1, 0.3]$.	In both subfigures , the left and right columns correspond to the independent case and correlated case respectively, and the top and bottom rows correspond to data generated from the model with homogeneous error and with heterogeneous error respectively. 

\begin{figure}[!ht]
	
	\centering
	\begin{subfigure}[!ht]{0.45\textwidth}
        \centering
        \includegraphics[width = \textwidth]{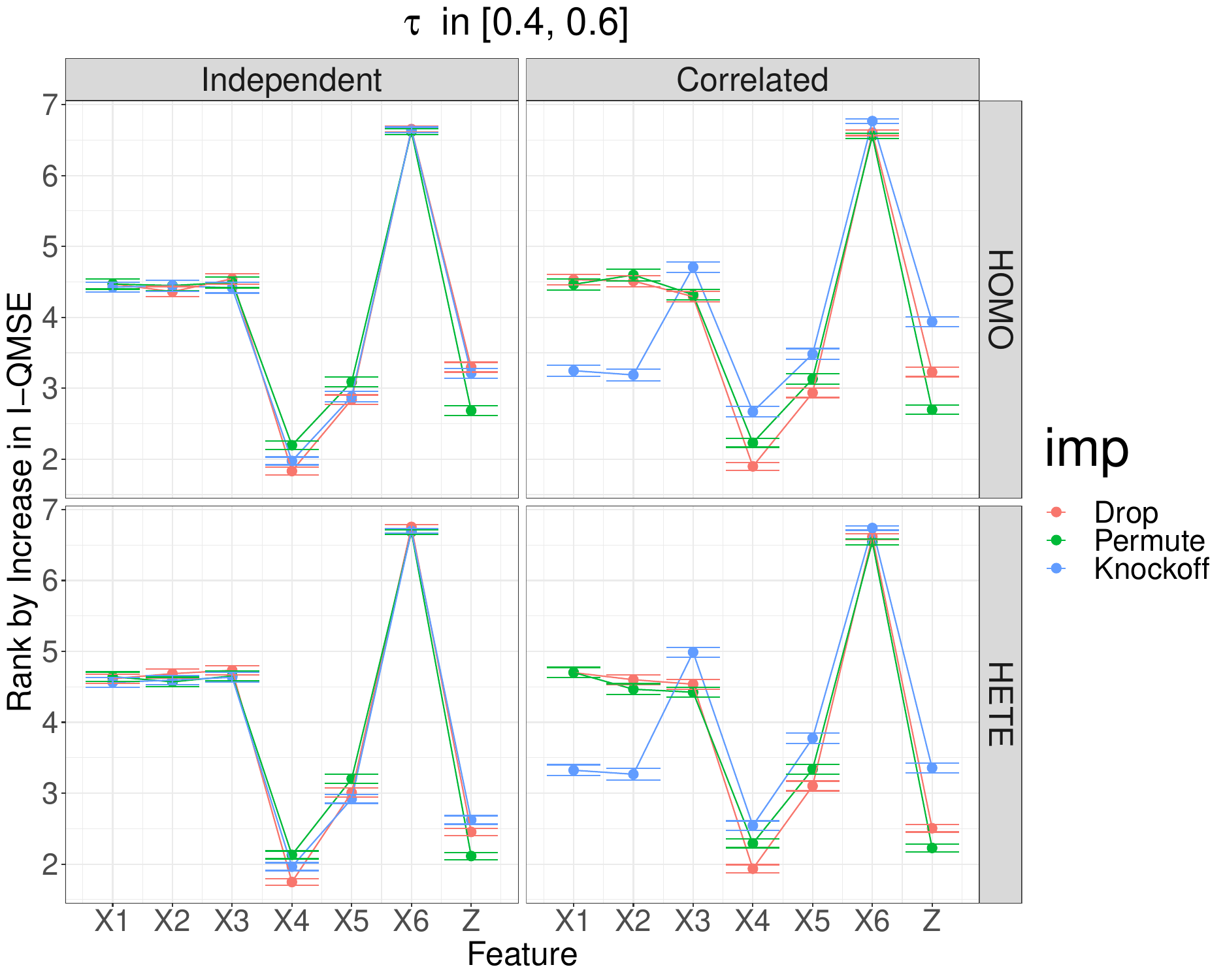}  
		\caption{$\tau \in [0.4, 0.6]$, $n = 100$ }
        \label{fig:vimp_rank_est_Tau40To60_n100}
    \end{subfigure}
	\hfill
    \begin{subfigure}[!ht]{0.45\textwidth}
        \centering
        \includegraphics[width = \textwidth]{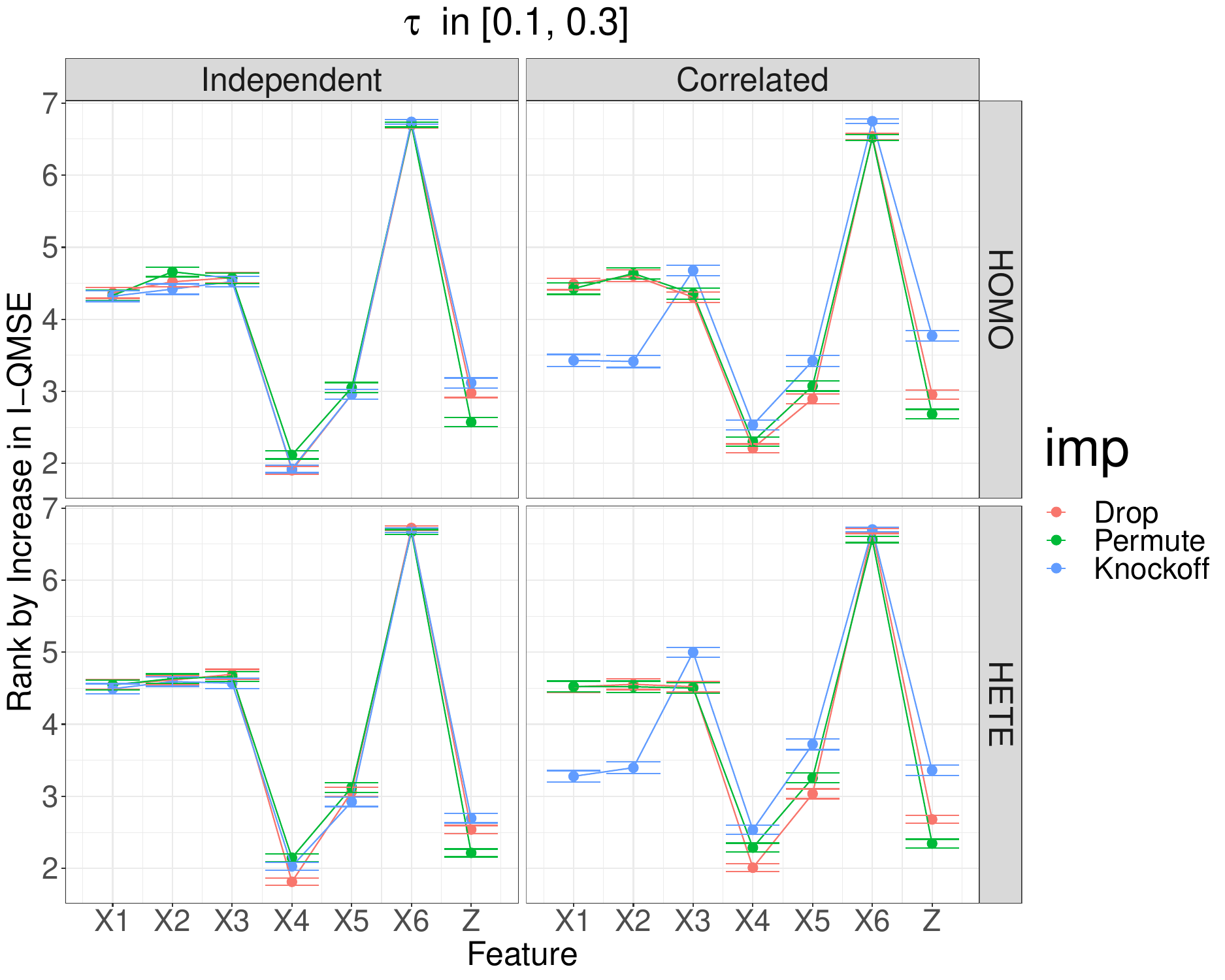}  
		\caption{$\tau \in [0.1, 0.3]$, $n = 100$ }
        \label{fig:vimp_rank_est_Tau10To30_n100}
    \end{subfigure}

    \begin{subfigure}[!ht]{0.45\textwidth}
        \centering
        \includegraphics[width = \textwidth]{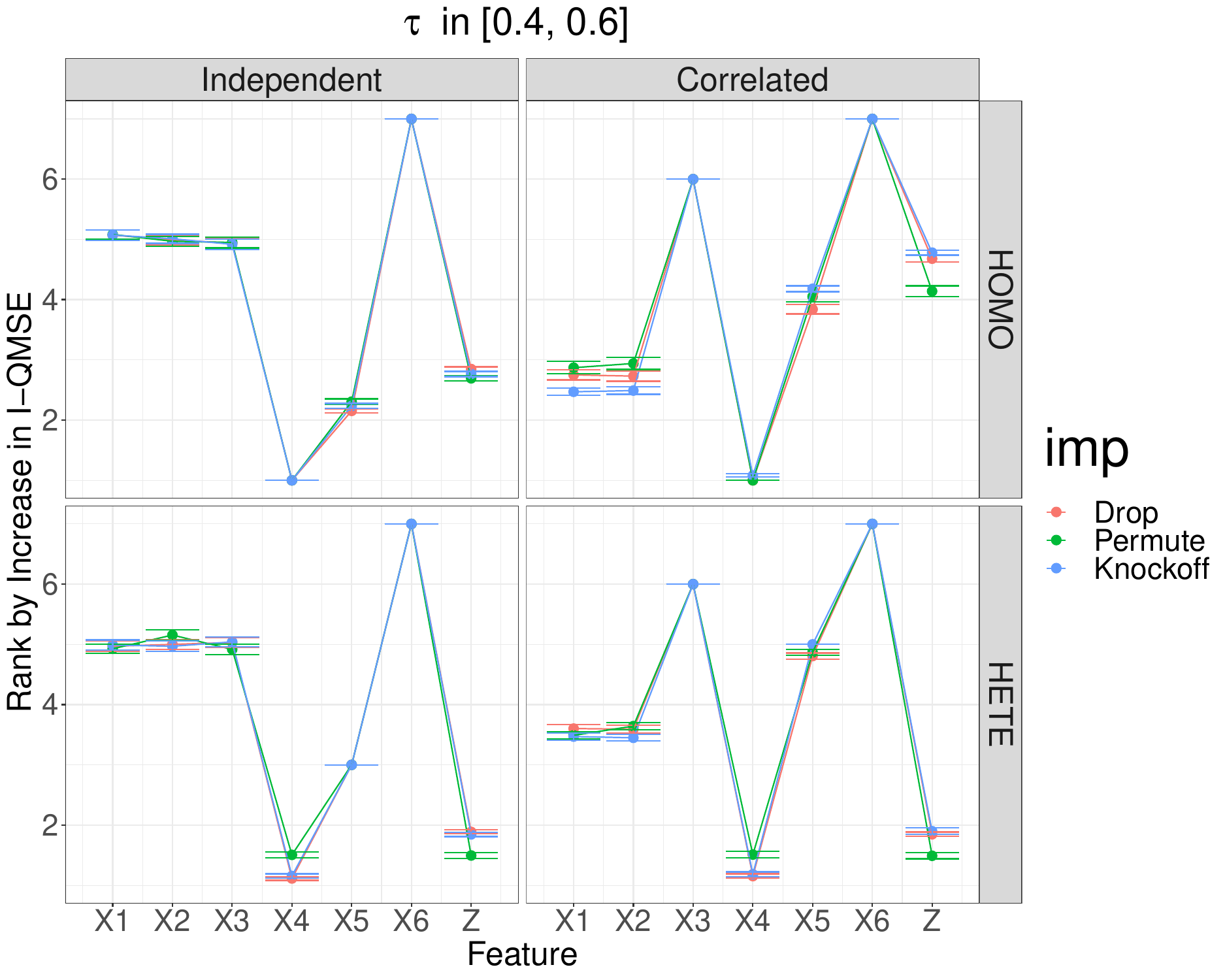}
		\caption{$\tau \in [0.4, 0.6]$, $n = 1000$ }
        \label{fig:vimp_rank_est_Tau40To60_n1000}
    \end{subfigure}
    \hfill
    \begin{subfigure}[!ht]{0.45\textwidth}
        \centering
        \includegraphics[width = \textwidth]{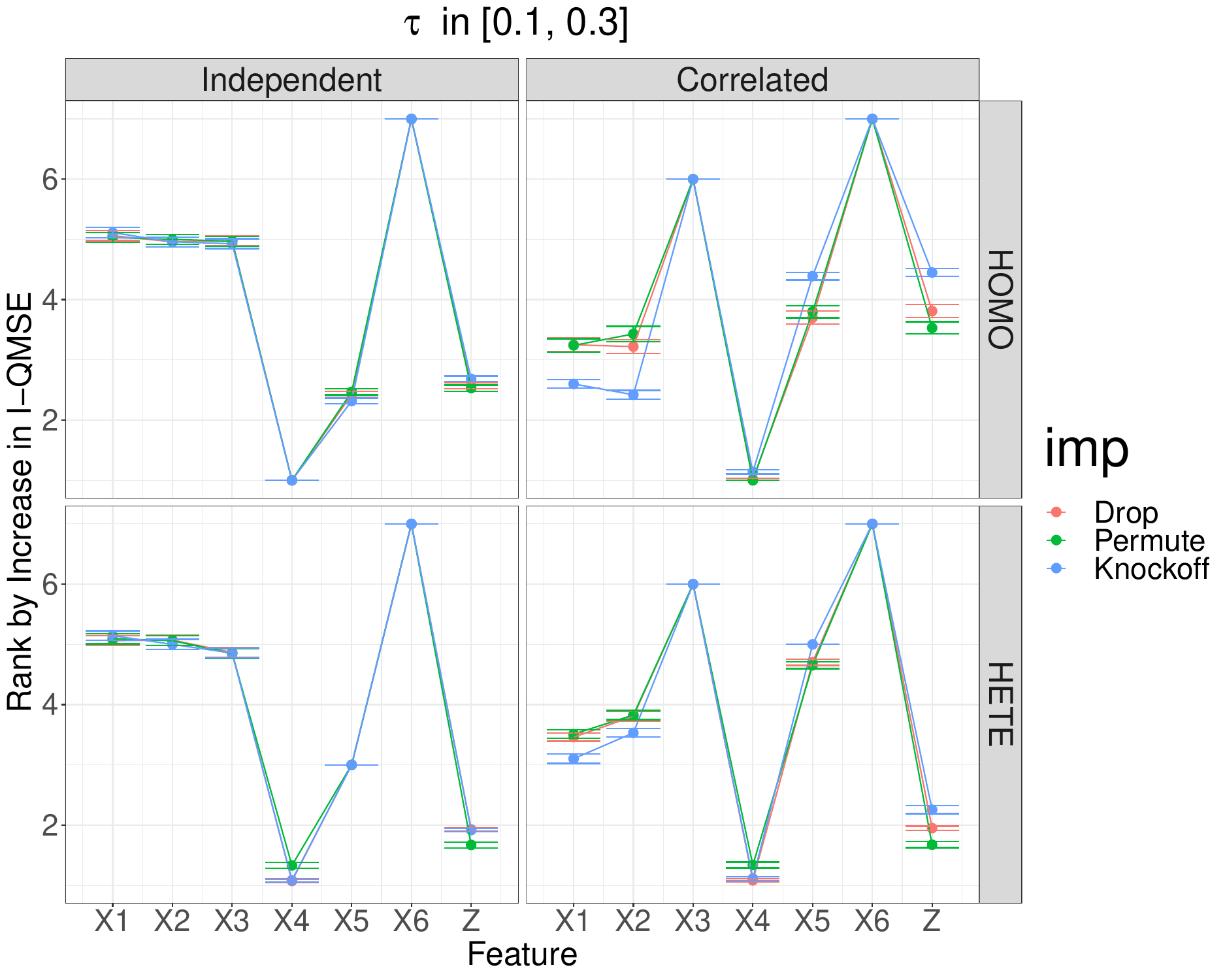}
		\caption{$\tau \in [0.1, 0.3]$, $n = 1000$ }
        \label{fig:vimp_rank_est_Tau10To30_n1000}
    \end{subfigure}
	\caption{\textit{Estimated} conditional feature importance ranking averaged over 100 simulations with training size $\bm{n=100}$ (top row) or $\bm{n=1000}$ (bottom row) based on increase in \textbf{I-QMSE} with $\tau \in \bm{[0.4, 0.6]}$ (left column) or $\tau \in \bm{[0.1, 0.3]}$ (right column). 
	Within each subfigure, the left and right columns correspond to the independent case ($\rho = 0$) and correlated case ($\rho = 0.9$) respectively with data generated from the model with homogeneous error (top row) or heterogeneous error (bottom row).
	In each setting, the three colors correspond to different ways of muting the effect of each feature of interest: dropping (green), permuting (green) and replacing with knockoff counterparts (blue). The error bar represents 1 standard error over 100 simulations.}
	\label{fig:vimp_rank_est_QMSE}
\end{figure}

We then carry out simulations with training size $n=100$ to illustrate the proposed feature importance ranking with 3-fold cross fitting. Simulations are repeated 100 times with the same tuning procedure as in Section \ref{sec:num_sim_predictive} and the average rankings based on increase in I-QMSE are plotted in Figure \ref{fig:vimp_rank_est_Tau40To60_n100} for $\tau \in [0.4, 0.6]$ and in Figure \ref{fig:vimp_rank_est_Tau10To30_n100} for $\tau \in [0.1, 0.3]$. The larger the rank, the more important the feature. In each figure, the colors corresponds to different ways muting the effect of each feature of interest: dropping (green), permuting (green) and replacing with knockoff counterparts (blue).

For $\tau \in [0.4, 0.6]$,  all 3 ways of muting effects can correctly rank the conditional importance of each feature in the independent HOMO case and independent HETE. However, the accuracy is less satisfying in the correlated case. Only estimated rankings based on replacements with knockoffs roughly follow the true rankings. For dropping and permuting, they correctly identify the most important one ($X_6$) and the least important one ($X_4$), but don't rank the rest features correctly. In particular, they don't capture the drop in conditional importance of $X_1$ and $X_2$ relative to $X_3$ due to correlation. Results for $\tau \in [0.1, 0.3]$ are similar.

Fortunately, the proposed feature importance measure improves with training size. Rerun simulations above with $n=1000$ and the results are plotted in Figure \ref{fig:vimp_rank_est_Tau40To60_n1000} for $\tau \in [0.4, 0.6]$ and in Figure \ref{fig:vimp_rank_est_Tau10To30_n1000} for $\tau \in [0.1, 0.3]$. In this case, estimated ranks of all features with any of the 3 ways of muting effects are the same as the true ranks in Figure \ref{fig:vimp_rank_true_Tau40To60} and Figure \ref{fig:vimp_rank_true_Tau10To30} except that $Z$ is ranked lower than the truth in the Correlated HETE case for $\tau \in [0.1, 0.3]$.

In practice, the performance can not be measured by I-QMSE but by I-QLoss instead. As plotted in the supplement material, conclusions based on I-QLoss are identical to those above based on I-QMSE.

Overall, simulations above suggest the proposed importance ranking measure should be promising in practice. As long as the sample size is moderately large, the proposed measure can capture the drop in conditional importance due to correlation and rank features entailing location-shift effects correctly. For features entailing scaling-shift effect, the proposed measure can still capture part of this effect but the resulting ranks may sometimes, for some quantile levels, lower than expected, especially relative to those entailing location-shift effect. We leave further investigations along this line as future work.

\subsection{Real Data}

Relationships between the response and features can be more complex for real data. Here we consider the \texttt{Boston Housing} data which has been widely utilized in literature as a benchmark for evaluating methods. It contains $n=506$ observations and $p=$12 features such as average number of rooms per dwelling (\texttt{rm}), proportion of lower status of the population (\texttt{lstat}), pupil-teacher ratio by town (\texttt{ptratio}), index of accessibility to radial highways (\texttt{rad}), full-value property-tax rate per \$10,000 (\texttt{tax}), per capita crime rate by town (\texttt{crim}),  weighted distances to five Boston employment centers (\texttt{dis}) and so on, and can be used to predict either the nitric oxides concentration or the median value of owner-occupied homes with the latter the focus in our analysis.
More detailed descriptions can be found at \url{https://www.cs.toronto.edu/~delve/data/boston/bostonDetail.html}. 

\begin{figure}[!hbt]
	\centering
	\includegraphics[width = 0.75\textwidth]{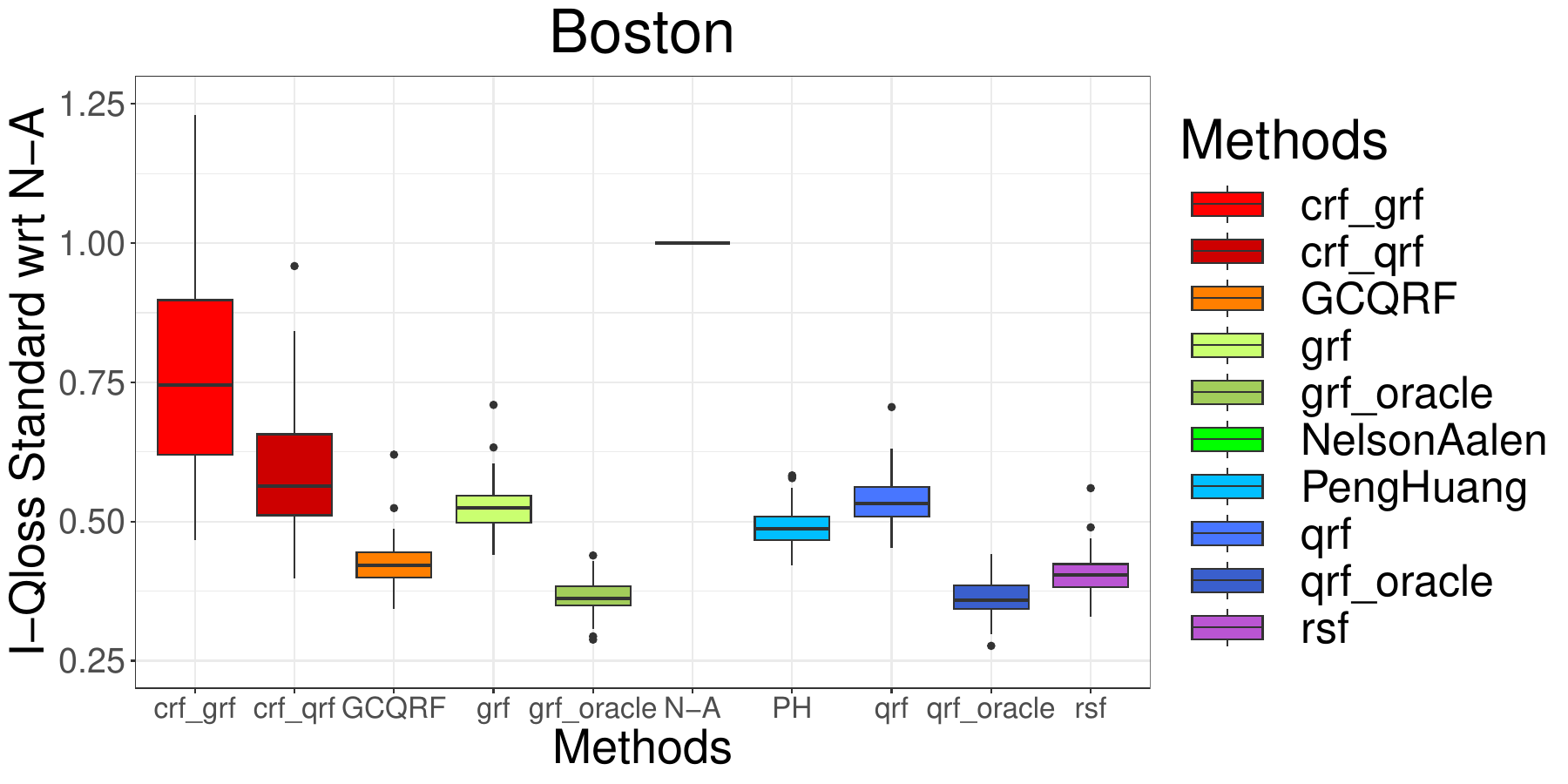}
	\includegraphics[width = 0.75\textwidth]{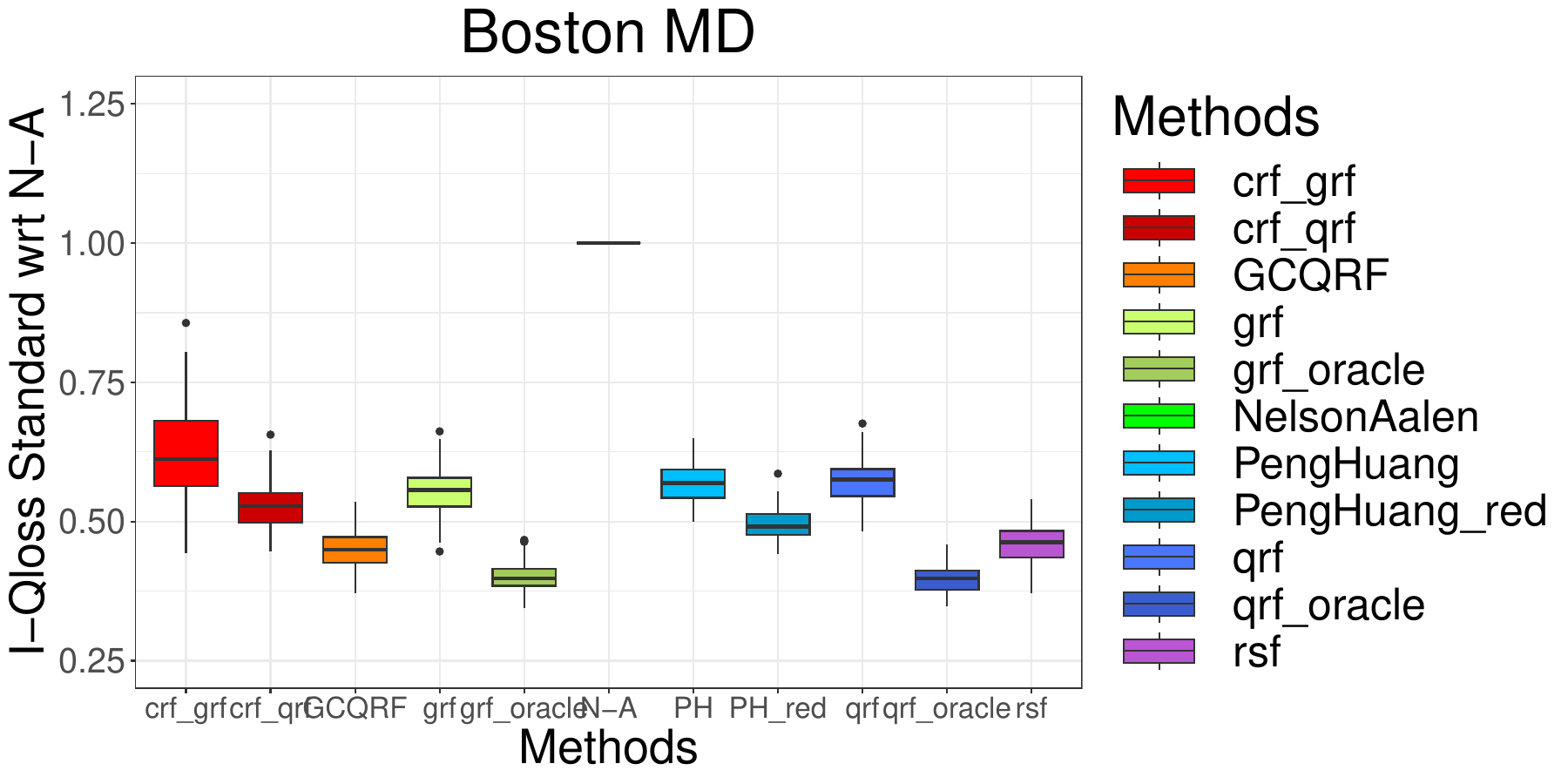}
	\includegraphics[width = 0.75\textwidth]{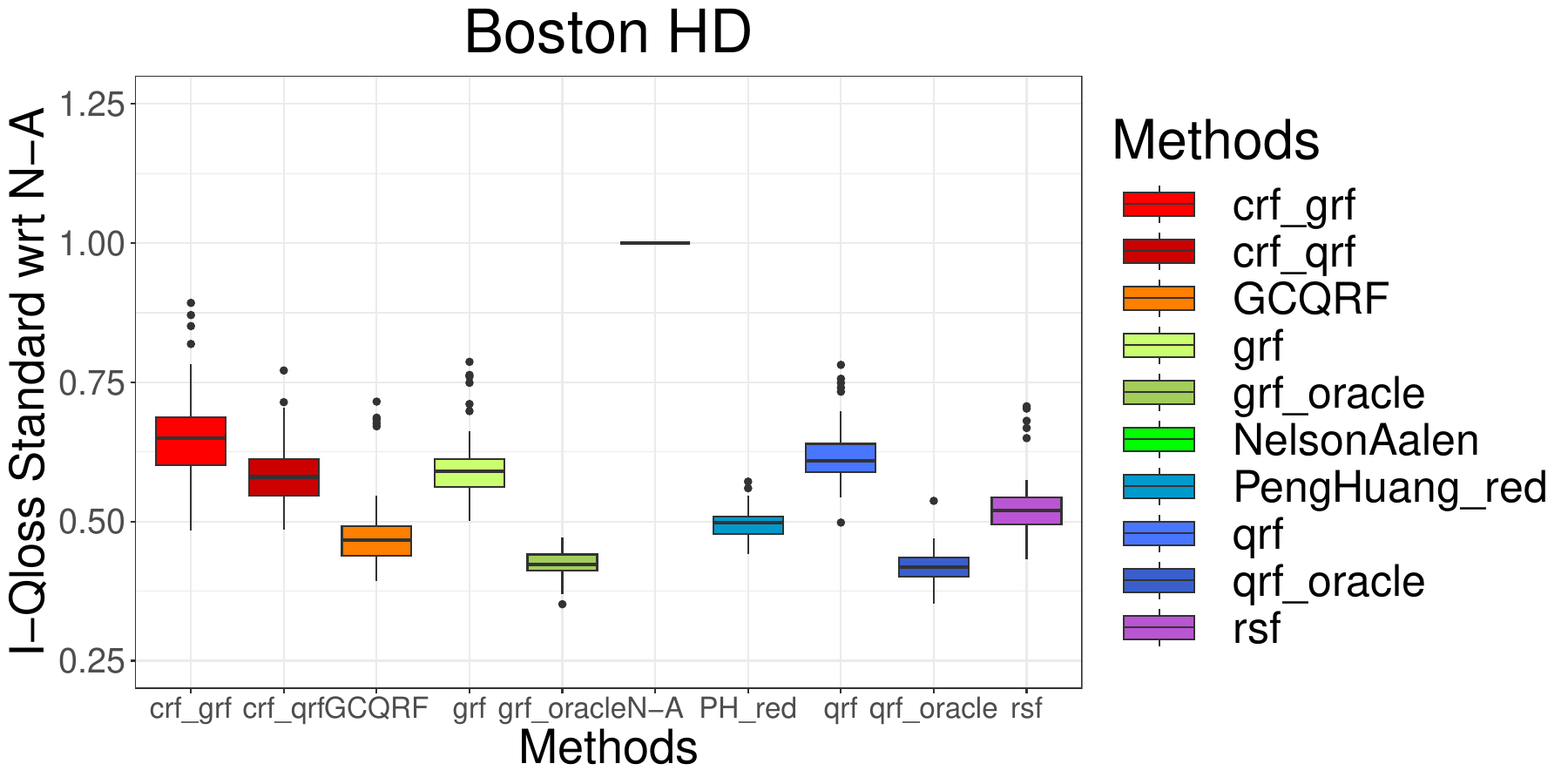}
	\caption{Comparison on predictive accuracy on the Boston Housing data}
	\label{fig:pfm_boston}
\end{figure}

Censoring is manually generated based on the categorical variable \texttt{rad}. For each level of \texttt{rad}, censoring is generated from a uniform distribution whose support is simulated such that the overall censoring rate of the entire data is around 25\%. With the true response known, we can compare performance based on the standard I-QLoss 
with grf-oracle and qrf-oracle included as references for forest-based approaches. 
On top of the original dataset, we add 50 and 500 noise variables generated from $N(0, \Sigma)$ with $\Sigma_{ij} = 0.3^{|i-j|}$ and call the resulting datasets \texttt{Boston MD} and \texttt{Boston HD} respectively. For these two datasets, we also consider PH-red, namely PH on the original 12 features only, as a reference.

To compare the predictive accuracy of GCQRF to alternative methods, for each dataset, randomly split it into two parts, 70\% for training and 30\% for testing. Parameters are tuned in the same way as with simulated data. The experiment is repeated 100 times and the boxplots of test standard I-QLoss with respect to the Nelson-Aalen estimator for a $\tau$-grid of $\{0.1, 0.15, \dots, 0.7\}$ are shown in Figure \ref{fig:pfm_boston}. The smaller the test standard I-QLoss, the higher the generalization accuracy for predicting conditional quantiles and the better the performance. Overall, consistent with results on the simulated data, GCQRF demonstrates competitive performance with high generalization accuracy and robustness to noise features.

For all three dataset, grf-oracle and qrf-oracle perform the best, as expected since no censoring adjustment is needed for these two. Other than the two oracle methods, GCQRF and rsf outperform the rest. Comparing GCQRF to rsf, rsf is slightly better than GCQRF while GCQRF is more robust than rsf as more noise features are added.

\begin{figure}[!hbt]
	\centering
		\centering
		\includegraphics[width = 0.45\textwidth]{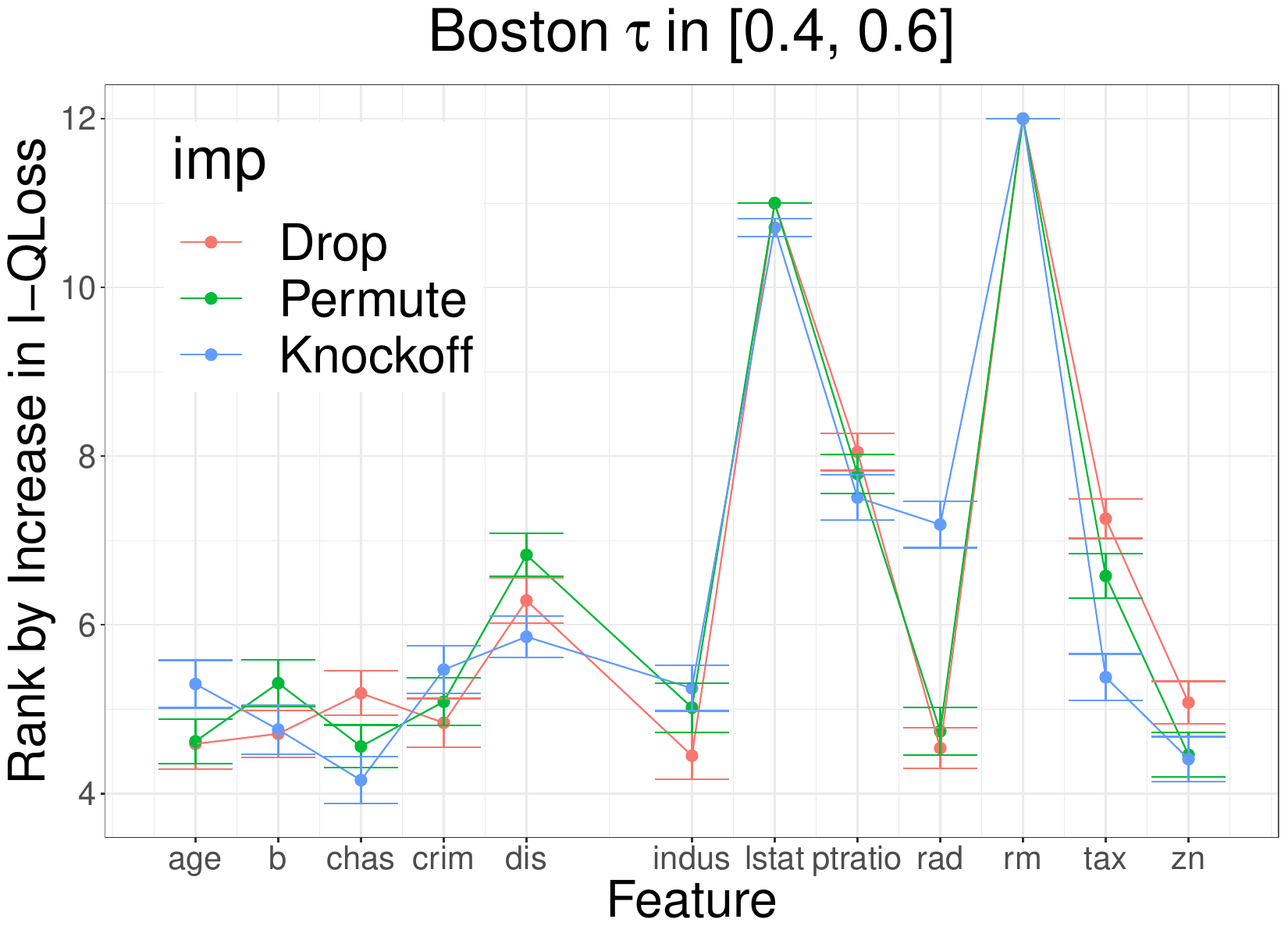}
		\includegraphics[width = 0.45\textwidth]{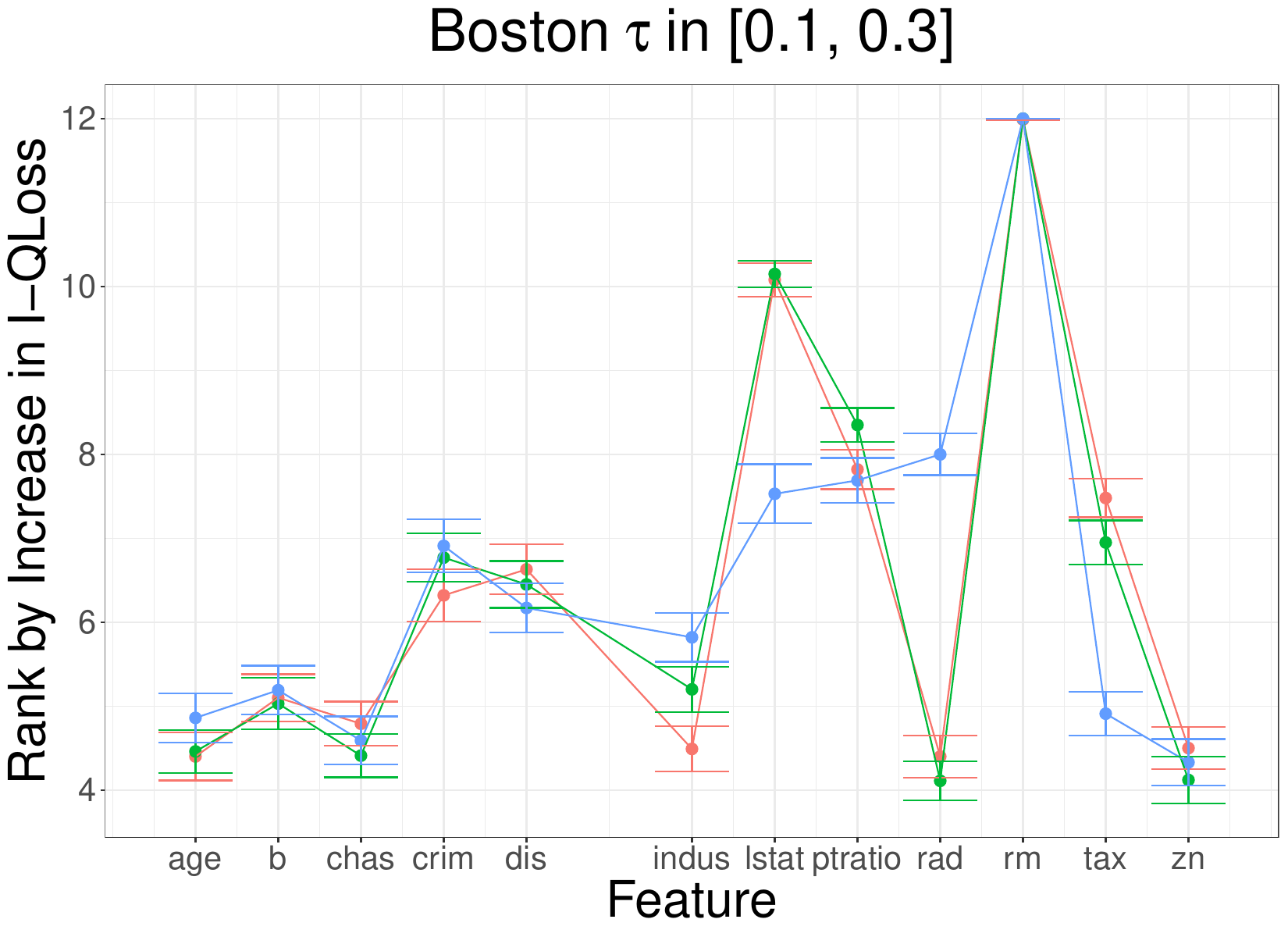}
		\caption{Estimated conditional feature importance ranking based on increase in I-QLoss for the Boston Housing data. Left: $\tau \in [0.4, 0.6]$; Right: $\tau \in [0.1, 0.3]$. The error bar represents 1 standard error across 100 repetitions.}
		\label{fig:vimp_boston}
\end{figure}

Next, we measure the conditional feature importance ranking of each feature with the proposed Algorithm \ref{alg:VIMP_Cross} on the original \texttt{Boston} data. The three different ways of modifying features, namely dropping, permuting and replacing with knockoffs, are all considered. 3-fold cross-fitting is utilized for stabilization. In addition, considering the randomness in growing GCQRFs, permuting features and resampling from knockoffs, we repeat the experiment 100 times. The average ranks based on increase in I-QLoss are plotted in Figure \ref{fig:vimp_boston}, the left one for $\tau \in [0.4, 0.6]$ and the right one for $\tau \in [0.1, 0.3]$. 

For $\tau \in [0.4, 0.6]$, the first five most importance features by GCQRF are 
\begin{itemize}
	\item Dropping: \texttt{rm}, \texttt{lstat}, \texttt{ptratio}, \texttt{tax}, \texttt{dis},
	\item Permuting: \texttt{rm}, \texttt{lstat}, \texttt{ptratio}, \texttt{tax}, \texttt{dis},
	\item Knockoffs: \texttt{rm}, \texttt{lstat}, \texttt{ptratio}, \texttt{rad}, \texttt{dis}.
\end{itemize}
Notice that the first three and the fifth most importance features are \texttt{rm}, \texttt{lstat}, \texttt{ptratio} and \texttt{dis} respectively, the same for dropping, permuting and replacing with knockoffs. The fourth most important one is \texttt{rad} by the knockoff approach but is \texttt{tax} by dropping and permuting.  In short, average number of rooms per dwelling (\texttt{rm}), proportion of lower status of the population (\texttt{lstat}), pupil-teacher ratio by town (\texttt{ptratio}) captures additional important information not reflected from the rest features on quantiles around the median of the median value of owner-occupied homes.

For $\tau \in [0.1, 0.3]$, ranks by GCQRF are mostly similar to those for $\tau \in [0.4, 0.6]$, 
except that
\begin{itemize}
	\item \texttt{crim} has a much higher rank, the fifth most important one, for the lower quantiles $\tau \in [0.1, 0.3]$ for all of dropping, permuting and replacing with knockoffs. Thus, the crime rate is estimated to play a more important role for lower quantile levels than for quantiles around the median.
	\item With the knockoff approach, \texttt{lstat} is ranked the fourth place, similar to \texttt{ptratio} and \texttt{rad} and lower than \texttt{rm}. In contrast, \texttt{lstat} is still ranked  the second most important by dropping and permuting.
\end{itemize}
This illustrates the heterogeneity of features effects on different quantiles of the median home value. In particular, the crime rate is more influential on low-valued home than on middle-valued one.

\section{Discussions}\label{sec:discussions}

This work proposed a novel random-forest-based approach for predicting conditional quantile processes for data potentially subject to right censoring. Our approach is flexible enough to work with nonlinear complex data, applicable for both low and high dimensional features, and rich of implicit regularization so as to adapt to data of various quality. Moreover, we seek to unravel a thorough and reliable understanding on the time-to-event by focusing on the conditional quantile process over an interval of quantiles levels rather than locally at some value. What's more, we propose a prediction-accuracy-based measure for ranking the importance of features, providing valuable insights for the relationship between features and the response under investigation.  Our numerical analysis indicates superior predictive accuracy of the proposed method over existing ones across various settings. One drawback of the proposed model is that it is required to re-estimate quantiles for each possible split, and therefore requires more computational power than the classical regression and classification random forests. A potential future work is to reduce the computational cost while maintaining the accuracy.

Theoretically, we reformulate the proposed GCQRF as a randomized incomplete IDUP and establish the corresponding weak convergence results by incorporating both randomized kernels and incomplete subsample selection. Our results suggest that the randomness of the tree kernel and of the selection of an incomplete subset of subsamples can result in substantially increased limiting variance, even dominating that in the complete case, and should be taken into account for better characterizing the variation of the predicted process.
These findings are in line with those in \cite{Peng2022} for real-valued predictions. 
A related interesting future work is to consistently estimate the variance function of the predicted process.

It is noticeable that our asymptotic analysis focuses on the variation of randomized IDUPs around the mean process of the kernel. The consistency of the tree and the forest can be established by requiring additional specifications on the tree structure, which have not been implemented in the proposed model, and is left as future work. In practice, such additional requirements may be seen as some sort of bias-variance trade-off. However, when predictive accuracy is the ultimate target, as in the case of accuracy-based variable importance measure, whether these requirements are beneficial is data-dependent.

\bibliographystyle{abbrvnat}
\bibliography{database}

\begin{thebibliography}{37}
\providecommand{\natexlab}[1]{#1}
\providecommand{\url}[1]{\texttt{#1}}
\expandafter\ifx\csname urlstyle\endcsname\relax
  \providecommand{\doi}[1]{doi: #1}\else
  \providecommand{\doi}{doi: \begingroup \urlstyle{rm}\Url}\fi

\bibitem[Athey et~al.(2019)Athey, Tibshirani, and Wager]{Athey2019}
S.~Athey, J.~Tibshirani, and S.~Wager.
\newblock Generalized random forests.
\newblock \emph{The Annals of Statistics}, 47\penalty0 (2):\penalty0
  1148--1178, 2019.

\bibitem[Biau and Scornet(2016)]{Biau2016}
G.~Biau and E.~Scornet.
\newblock A random forest guided tour.
\newblock \emph{Test}, 25:\penalty0 197--227, 2016.

\bibitem[Breiman(2001)]{Breiman2001}
L.~Breiman.
\newblock Random forests.
\newblock \emph{Machine learning}, 45:\penalty0 5--32, 2001.

\bibitem[Candes et~al.(2018)Candes, Fan, Janson, and Lv]{Candes2018}
E.~Candes, Y.~Fan, L.~Janson, and J.~Lv.
\newblock Panning for gold:‘model-x’knockoffs for high dimensional
  controlled variable selection.
\newblock \emph{Journal of the Royal Statistical Society Series B: Statistical
  Methodology}, 80\penalty0 (3):\penalty0 551--577, 2018.

\bibitem[Chen and Kato(2019)]{Chen2019}
X.~Chen and K.~Kato.
\newblock {Randomized incomplete $U$-statistics in high dimensions}.
\newblock \emph{The Annals of Statistics}, 47\penalty0 (6):\penalty0 3127 --
  3156, 2019.

\bibitem[Cho et~al.(2022{\natexlab{a}})Cho, Holloway, Couper, and
  Kosorok]{Cho2023}
H.~Cho, S.~T. Holloway, D.~J. Couper, and M.~R. Kosorok.
\newblock {Multi-stage optimal dynamic treatment regimes for survival outcomes
  with dependent censoring}.
\newblock \emph{Biometrika}, 110\penalty0 (2):\penalty0 395--410, 08
  2022{\natexlab{a}}.

\bibitem[Cho et~al.(2022{\natexlab{b}})Cho, Jewell, and Kosorok]{Cho2022}
H.~Cho, N.~P. Jewell, and M.~R. Kosorok.
\newblock Interval censored recursive forests.
\newblock \emph{Journal of Computational and Graphical Statistics}, 31\penalty0
  (2):\penalty0 390--402, 2022{\natexlab{b}}.

\bibitem[Christou and Akritas(2019)]{Christou2019}
E.~Christou and M.~G. Akritas.
\newblock Single index quantile regression for censored data.
\newblock \emph{Statistical Methods \& Applications}, 28\penalty0 (4):\penalty0
  655--678, 2019.

\bibitem[Fern{\'a}ndez-Delgado et~al.(2014)Fern{\'a}ndez-Delgado, Cernadas,
  Barro, and Amorim]{Fernandez2014}
M.~Fern{\'a}ndez-Delgado, E.~Cernadas, S.~Barro, and D.~Amorim.
\newblock Do we need hundreds of classifiers to solve real world classification
  problems?
\newblock \emph{The journal of machine learning research}, 15\penalty0
  (1):\penalty0 3133--3181, 2014.

\bibitem[Hastie et~al.(2009)Hastie, Tibshirani, Friedman, and Friedman]{ESL}
T.~Hastie, R.~Tibshirani, J.~H. Friedman, and J.~H. Friedman.
\newblock \emph{The elements of statistical learning: data mining, inference,
  and prediction}, volume~2.
\newblock Springer, 2009.

\bibitem[Hastie et~al.(2020)Hastie, Tibshirani, and Tibshirani]{Hastie2020}
T.~Hastie, R.~Tibshirani, and R.~Tibshirani.
\newblock {Best Subset, Forward Stepwise or Lasso? Analysis and Recommendations
  Based on Extensive Comparisons}.
\newblock \emph{Statistical Science}, 35\penalty0 (4):\penalty0 579 -- 592,
  2020.

\bibitem[Heilig and Nolan(2001)]{Heilig2001}
C.~Heilig and D.~Nolan.
\newblock Limit theorems for the infinite-degree u-process.
\newblock \emph{Statistica Sinica}, pages 289--302, 2001.

\bibitem[Heilig(1997)]{Heilig1997}
C.~M. Heilig.
\newblock \emph{An empirical process approach to U-processes of increasing
  degree}.
\newblock University of California, Berkeley, 1997.

\bibitem[Hooker et~al.(2021)Hooker, Mentch, and Zhou]{Hooker2021}
G.~Hooker, L.~Mentch, and S.~Zhou.
\newblock Unrestricted permutation forces extrapolation: variable importance
  requires at least one more model, or there is no free variable importance.
\newblock \emph{Statistics and Computing}, 31:\penalty0 1--16, 2021.

\bibitem[Hothorn et~al.(2004)Hothorn, Lausen, Benner, and
  Radespiel-Tr{\"o}ger]{Hothorn2004}
T.~Hothorn, B.~Lausen, A.~Benner, and M.~Radespiel-Tr{\"o}ger.
\newblock Bagging survival trees.
\newblock \emph{Statistics in medicine}, 23\penalty0 (1):\penalty0 77--91,
  2004.

\bibitem[Ishwaran et~al.(2008)Ishwaran, Kogalur, Blackstone, and
  Lauer]{Ishwaran2008}
H.~Ishwaran, U.~B. Kogalur, E.~H. Blackstone, and M.~S. Lauer.
\newblock Random survival forests.
\newblock \emph{The annals of applied statistics}, 2\penalty0 (3):\penalty0
  841--860, 2008.

\bibitem[Lee(1990)]{uStat}
A.~J. Lee.
\newblock \emph{U-statistics : Theory and Practice.}
\newblock CRC Press, 1990.

\bibitem[Li and Bradic(2020)]{Li2020}
A.~H. Li and J.~Bradic.
\newblock Censored quantile regression forest.
\newblock In \emph{International Conference on Artificial Intelligence and
  Statistics}, pages 2109--2119. PMLR, 2020.

\bibitem[Li and Peng(2017)]{Li2017}
R.~Li and L.~Peng.
\newblock Assessing quantile prediction with censored quantile regression
  models.
\newblock \emph{Biometrics}, 73\penalty0 (2):\penalty0 517--528, 2017.

\bibitem[Meinshausen(2006)]{Meinshausen2006}
N.~Meinshausen.
\newblock Quantile regression forests.
\newblock \emph{Journal of machine learning research}, 7, 2006.

\bibitem[Mentch and Hooker(2016)]{Mentch2016}
L.~Mentch and G.~Hooker.
\newblock Quantifying uncertainty in random forests via confidence intervals
  and hypothesis tests.
\newblock \emph{Journal of Machine Learning Research}, 17\penalty0
  (26):\penalty0 1--41, 2016.

\bibitem[Mentch and Zhou(2020)]{Mentch2020}
L.~Mentch and S.~Zhou.
\newblock Randomization as regularization: A degrees of freedom explanation for
  random forest success.
\newblock \emph{The Journal of Machine Learning Research}, 21\penalty0
  (1):\penalty0 6918--6953, 2020.

\bibitem[Mentch and Zhou(2022)]{Mentch2022}
L.~Mentch and S.~Zhou.
\newblock Getting better from worse: Augmented bagging and a cautionary tale of
  variable importance.
\newblock \emph{The Journal of Machine Learning Research}, 23\penalty0
  (1):\penalty0 10157--10188, 2022.

\bibitem[Peng and Huang(2008)]{Peng2008}
L.~Peng and Y.~Huang.
\newblock Survival analysis with quantile regression models.
\newblock \emph{Journal of the American Statistical Association}, 103\penalty0
  (482):\penalty0 637--649, 2008.

\bibitem[Peng et~al.(2022)Peng, Coleman, and Mentch]{Peng2022}
W.~Peng, T.~Coleman, and L.~Mentch.
\newblock Rates of convergence for random forests via generalized u-statistics.
\newblock \emph{Electronic Journal of Statistics}, 16\penalty0 (1):\penalty0
  232--292, 2022.

\bibitem[Pollard(1984)]{Pollard1984}
D.~Pollard.
\newblock \emph{Convergence of Stochastic Processes}.
\newblock Springer New York, NY, 1984.

\bibitem[Pollard(1990)]{Pollard1990}
D.~Pollard.
\newblock Empirical processes: theory and applications.
\newblock Institute of Mathematical Statistics, 1990.

\bibitem[Probst et~al.(2019)Probst, Wright, and Boulesteix]{Probst2019}
P.~Probst, M.~N. Wright, and A.-L. Boulesteix.
\newblock Hyperparameters and tuning strategies for random forest.
\newblock \emph{Wiley Interdisciplinary Reviews: data mining and knowledge
  discovery}, 9\penalty0 (3):\penalty0 e1301, 2019.

\bibitem[Scornet(2017)]{Scornet2017}
E.~Scornet.
\newblock Tuning parameters in random forests.
\newblock \emph{ESAIM: Proceedings and Surveys}, 60:\penalty0 144--162, 2017.

\bibitem[Scornet et~al.(2015)Scornet, Biau, and Vert]{Scornet2015}
E.~Scornet, G.~Biau, and J.-P. Vert.
\newblock {Consistency of random forests}.
\newblock \emph{The Annals of Statistics}, 43\penalty0 (4):\penalty0 1716 --
  1741, 2015.

\bibitem[Van Der~Vaart and Wellner(1996)]{Van1996}
A.~W. Van Der~Vaart and J.~A. Wellner.
\newblock \emph{Weak convergence}.
\newblock Springer, 1996.

\bibitem[Wager and Athey(2018)]{Wager2018}
S.~Wager and S.~Athey.
\newblock Estimation and inference of heterogeneous treatment effects using
  random forests.
\newblock \emph{Journal of the American Statistical Association}, 113\penalty0
  (523):\penalty0 1228--1242, 2018.

\bibitem[Wang and Wang(2009)]{Wang2009}
H.~J. Wang and L.~Wang.
\newblock Locally weighted censored quantile regression.
\newblock \emph{Journal of the American Statistical Association}, 104\penalty0
  (487):\penalty0 1117--1128, 2009.

\bibitem[Williamson et~al.(2021)Williamson, Gilbert, Simon, and
  Carone]{Williamson2021}
B.~D. Williamson, P.~B. Gilbert, N.~R. Simon, and M.~Carone.
\newblock A general framework for inference on algorithm-agnostic variable
  importance.
\newblock \emph{Journal of the American Statistical Association}, pages 1--14,
  2021.

\bibitem[Zheng et~al.(2018)Zheng, Peng, and He]{Zheng2018}
Q.~Zheng, L.~Peng, and X.~He.
\newblock {High dimensional censored quantile regression}.
\newblock \emph{The Annals of Statistics}, 46\penalty0 (1):\penalty0 308 --
  343, 2018.

\bibitem[Zhou and Mentch(2023)]{Zhou2023}
S.~Zhou and L.~Mentch.
\newblock Trees, forests, chickens, and eggs: when and why to prune trees in a
  random forest.
\newblock \emph{Statistical Analysis and Data Mining: The ASA Data Science
  Journal}, 16\penalty0 (1):\penalty0 45--64, 2023.

\bibitem[Zhu et~al.(2022)Zhu, Sun, and Wei]{Zhu2022}
H.~Zhu, Y.~Sun, and Y.~Wei.
\newblock Hybrid censored quantile regression forest to assess the
  heterogeneous effects.
\newblock \emph{arXiv preprint arXiv:2212.05672}, 2022.

\end{thebibliography}

\end{document}